\pdfoutput=1 
\documentclass[runningheads]{llncs}
\usepackage{graphicx}

\usepackage{tikz}
\usepackage{comment} 
\usepackage{amsmath,amssymb} 
\usepackage{color}

\usepackage{enumitem}

\usepackage{tikz}
\usepackage{pgfplots}
\usetikzlibrary{pgfplots.groupplots}
\usepgfplotslibrary{groupplots}

\usepackage{booktabs}       
\usepackage{tabularx}

\usepackage{fp}

\usepackage{url}

\newcommand{\etal}{\textit{et al}.}
\newcommand{\ie}{\textit{i}.\textit{e}.}
\newcommand{\eg}{\textit{e}.\textit{g}.}
\newcommand\sbullet[1][.5]{\mathbin{\vcenter{\hbox{\scalebox{#1}{$\bullet$}}}}}

\graphicspath{{figures/}{figures/auto/}}

\begin{document}
	\pagestyle{headings}
	\mainmatter
	\def\ECCVSubNumber{2883}  
	
	\title{A Generic Visualization Approach for Convolutional Neural Networks} 

	\titlerunning{L2-Norm Constrained Attention Filter (L2-CAF)}
	%
	\author{Ahmed Taha
		\and		Xitong Yang		\and		Abhinav Shrivastava		\and 		Larry Davis
	}
	\institute{University of Maryland, College Park}
	\authorrunning{A. Taha \textit{et al.}}
	\maketitle
	
	\begin{abstract}
	
	Retrieval networks are essential for searching and indexing.
	Compared to classification networks, attention visualization for retrieval networks is hardly studied.  
	We formulate attention visualization as a constrained optimization problem. We leverage the unit \underline{L2}-Norm \underline{c}onstraint as an \underline{a}ttention \underline{f}ilter (L2-CAF) to localize attention in both classification and retrieval networks. Unlike recent literature, our approach requires neither architectural changes nor fine-tuning. Thus, a pre-trained network's performance is never undermined
	
	L2-CAF is quantitatively evaluated using weakly supervised object localization. State-of-the-art results are achieved on classification networks. For retrieval networks, significant improvement margins are achieved over a Grad-CAM baseline. Qualitative evaluation demonstrates how the L2-CAF visualizes attention per frame for a recurrent retrieval network. Further ablation studies highlight the computational cost of our approach and compare L2-CAF with other feasible alternatives. Code available  at
	\textit{https://bit.ly/3iDBLFv}

	  
\end{abstract}
	\section{Introduction}

Both classification and retrieval neural networks need attention visualization tools. These tools are important in medical and autonomous navigation to understand and interpret networks' decisions. Moreover, attention visualization enables weakly supervised object localization (WSOL) which reduces the cost of data annotation. WSOL avoids bounding-box labeling required by fully supervised approaches. Attention visualization and WSOL have been intensively studied for classification architectures~\cite{zeiler2014visualizing,zhou2016learning,singh2017hide,selvaraju2017grad,zhang2018adversarial,choe2019attention}. However, these approaches do not address retrieval networks.
In this paper, we leverage the unit L2-Norm constraint as an attention filter (L2-CAF) that works for both classification and retrieval neural networks, as shown in Figure~\ref{fig:l2_variants}.

\newcommand{\VaraintsImgSize}{0.15}
\begin{figure}[t]
	\scriptsize
	\centering
	\setlength\tabcolsep{1.0pt} 
	\renewcommand{\arraystretch}{1.0}	
	\begin{tabular}{@{}cccc c cccc@{}}
		Class-oblivious &\phantom{}& \multicolumn{2}{c}{Class-specific} & \phantom{abc}& Class-oblivious &\phantom{}& \multicolumn{2}{c}{Class-specific} 
		\\ 
		\includegraphics[width=\VaraintsImgSize\textwidth,height=\VaraintsImgSize\textwidth]{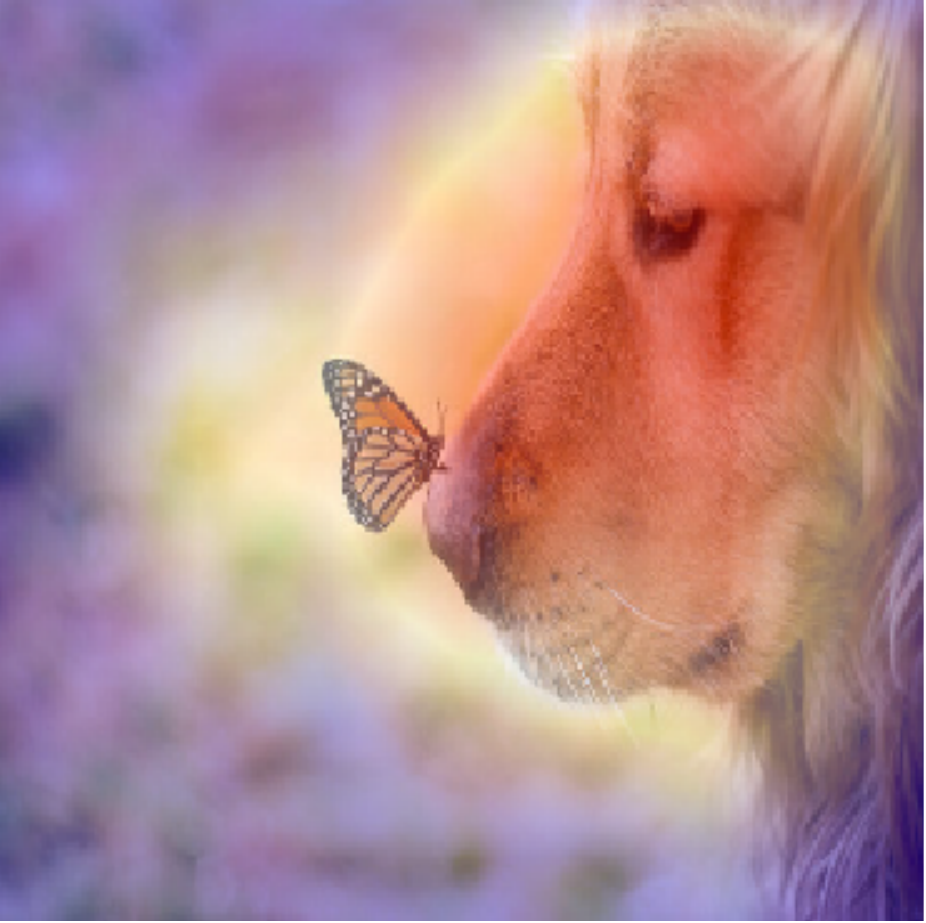} &&
		\includegraphics[width=\VaraintsImgSize\textwidth,height=\VaraintsImgSize\textwidth]{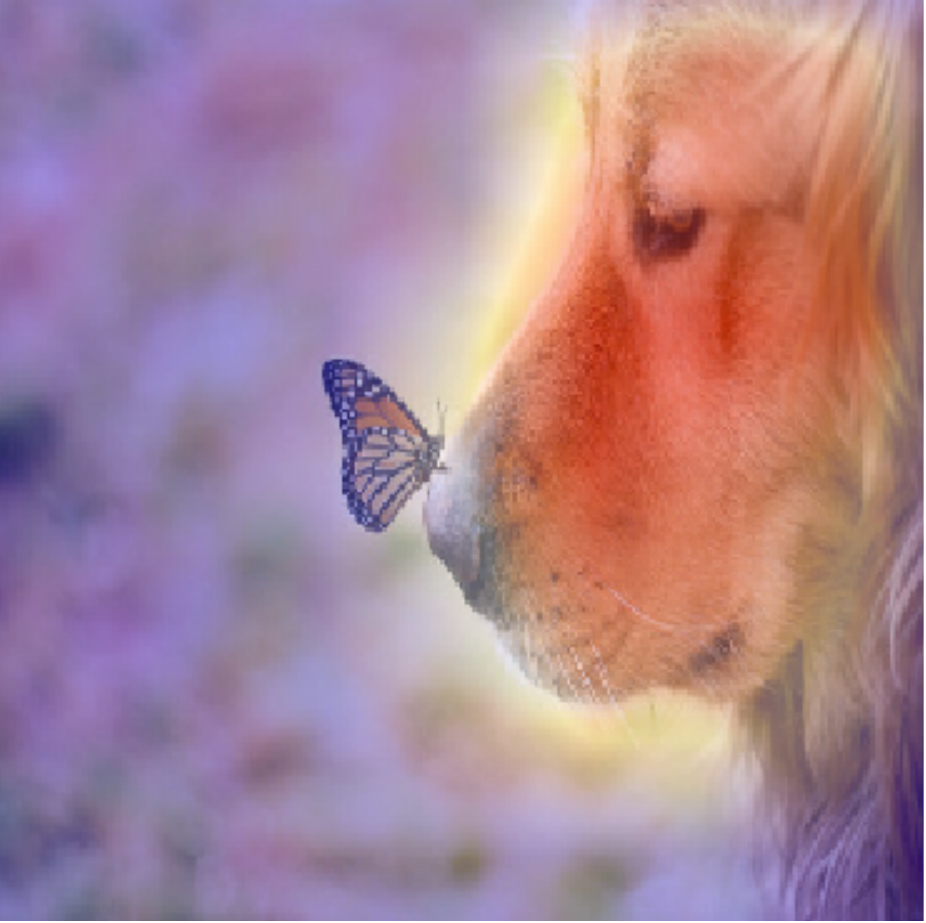} & \includegraphics[width=\VaraintsImgSize\textwidth,height=\VaraintsImgSize\textwidth]{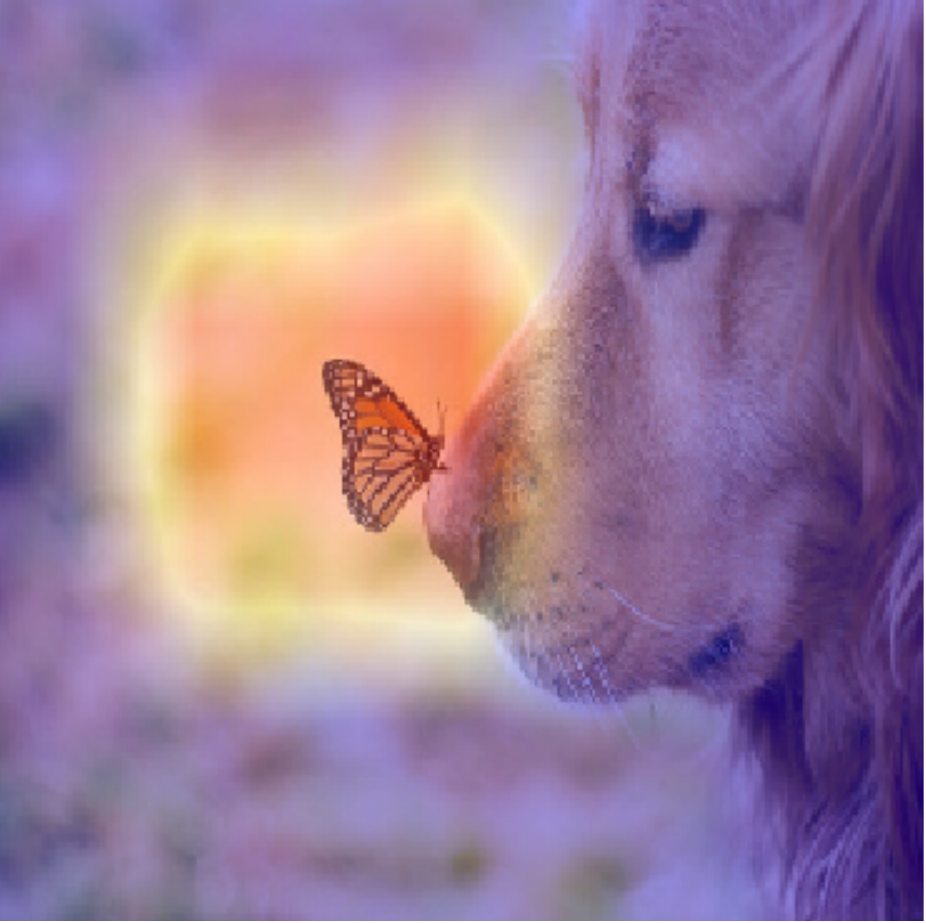} &&
		
		\includegraphics[width=\VaraintsImgSize\textwidth,height=\VaraintsImgSize\textwidth]{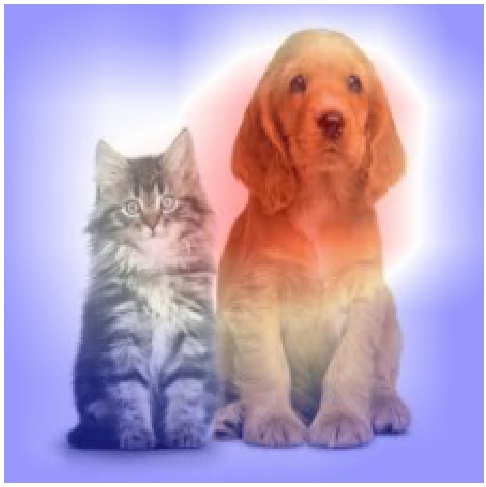} &&
		\includegraphics[width=\VaraintsImgSize\textwidth,height=\VaraintsImgSize\textwidth]{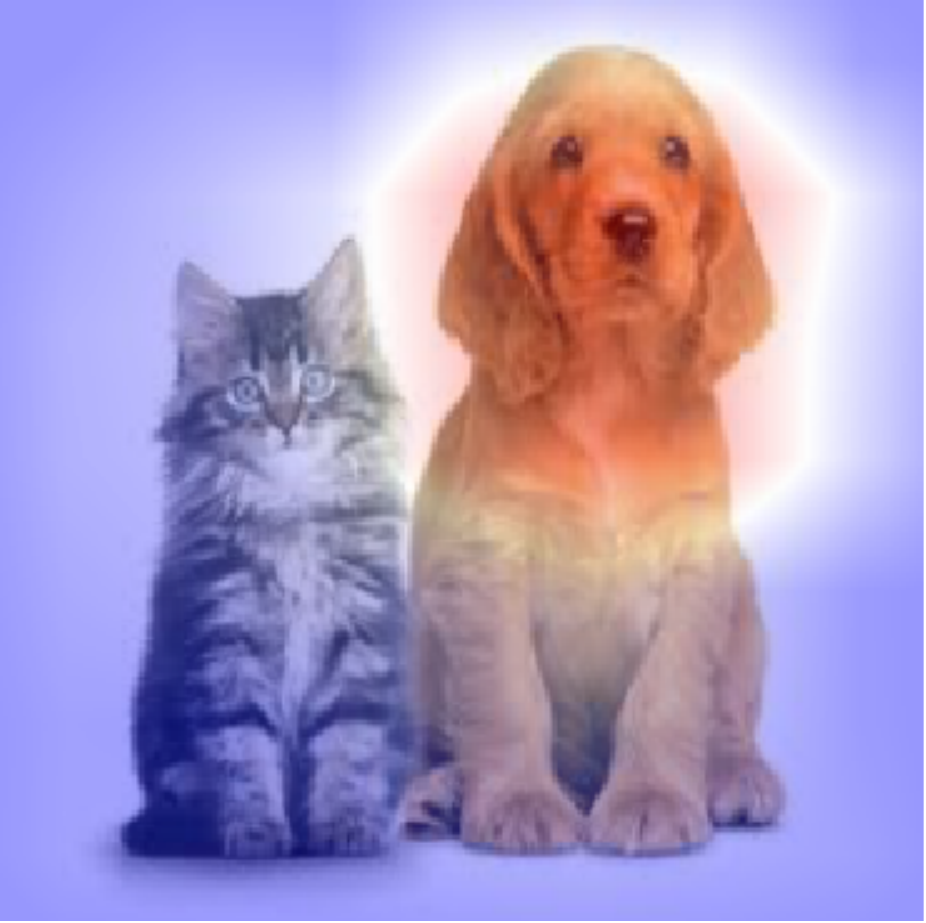} & \includegraphics[width=\VaraintsImgSize\textwidth,height=\VaraintsImgSize\textwidth]{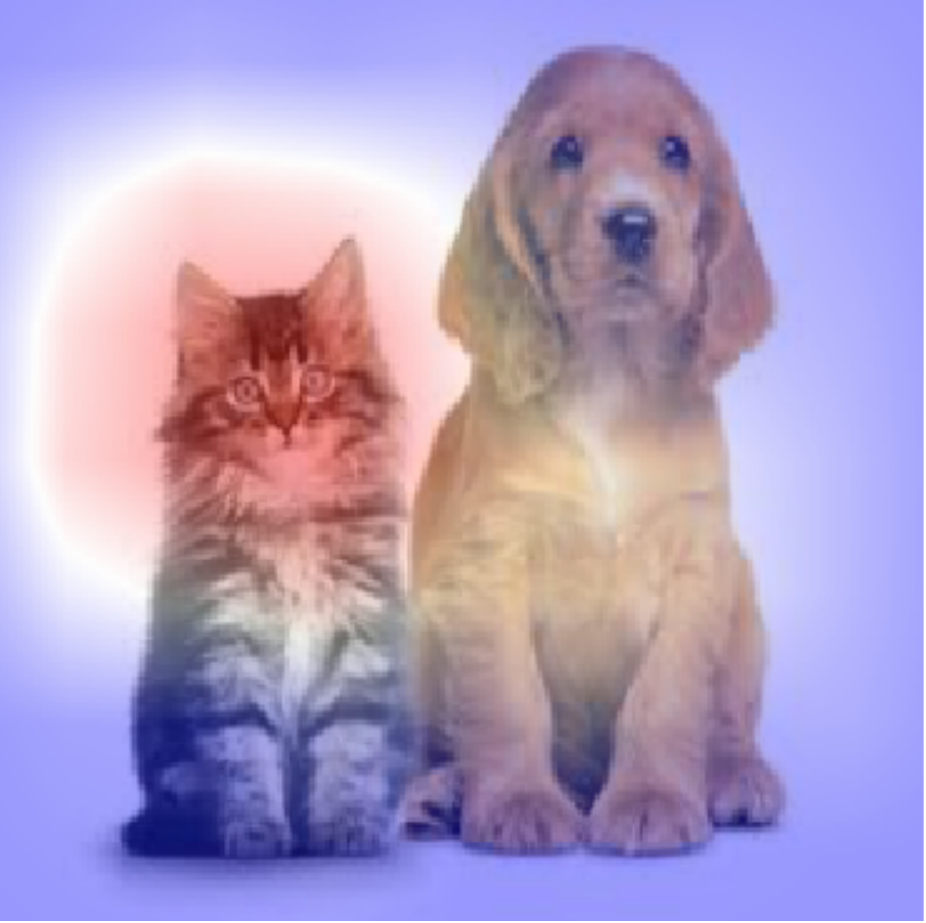} \\
	\end{tabular}
	\caption{L2-CAF enables both class-oblivious and class-specific visualizations. This separates our work from dominant literature that targets classification networks only. The supplementary video shows more vivid and challenging visualizations.}
	\label{fig:l2_variants}
\end{figure}

For classification networks, Zhou~\etal~\cite{zhou2016learning} propose class activation maps (CAM) for attention visualization and WSOL. Further research~\cite{singh2017hide,zhang2018adversarial,zhang2018self,choe2019attention} improved WSOL by  augmenting the most discriminative region with other less discriminative parts,~\eg, augment a cat's head with its legs. This improvement comes at the cost of few drawbacks: (1) They impose architectural constraints,~\eg, global average pooling (GAP) layer; (2) While fine-tuning boosts localization efficiency, it degrades classification accuracy. Grad-CAM~\cite{selvaraju2017grad} avoids these limitations, but it is originally formulated for classification networks.





\begin{figure*}[t]
\centering
\scriptsize
\includegraphics[width=1.0\textwidth]{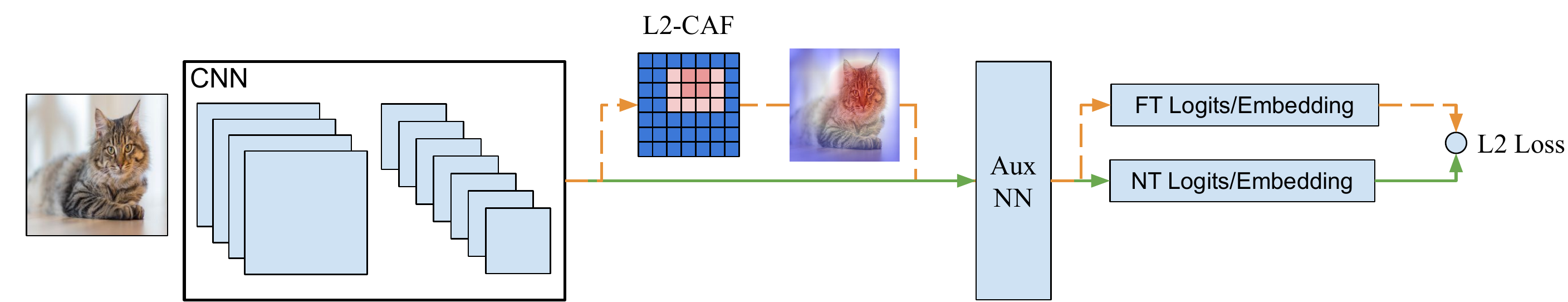}
\caption{An overview of the proposed unit L2-Norm constrained attention filter (L2-CAF). Given a pre-trained CNN with an auxiliary head (Aux NN), feed an input frame through a normal feed-forward pass (green solid path) to generate the network output logits/embedding $NT(x)$. Then, feed the same input again but multiply the last convolutional feature map by a constrained attention filter $f$ (orange dashed path) to generate a new filtered output $FT(x,f)$. Optimize the filter's weights through gradient descent to minimize the difference between $NT(x)$ and $FT(x, f)$. In standard CNN architectures, the L2-CAF is typically $7\times7$,~\ie, a cheap optimization problem $\in R^{49}$.}
\label{fig:overview}
\end{figure*}

Retrieval networks are essential for visual search~\cite{oh2017deep,kim2019deep}, zero-shot learning~\cite{bucher2016improving,yuan2017hard,zhang2016zero}, and fine-grained retrieval~\cite{sohn2016improved,oh2016deep}.  The large metric learning~\cite{oh2016deep,wang2017deep,chen2017beyond} and product quantization~\cite{cao2016deep,li2017deep,eghbali2019deep} literature reflect their importance. Despite that, attention visualization for retrieval networks has not been evaluated quantitatively. It is more challenging compared to classification due to the network's output -- a class-oblivious embedding.

The \underline{main contribution} of this paper is to leverage the L2-CAF as a visualization filter to identify key features of both classification and retrieval networks' output. Figure~\ref{fig:overview} illustrates the approach. Given a pre-trained CNN, feeding the same input $x$ through the network (green solid path) will always generate the same output $NT(x)$. If the final convolutional feature map is multiplied by a constrained attention filter $f$ in an element-wise manner (orange dashed path), the network generates a filtered output $FT(x, f)$.  Through gradient descent, we optimize $f$ to minimize the L2 loss  $L = ||NT(x)-FT(x, f)||^2$. The optimized filter $f$ reveals key spatial regions,~\eg, the cat's head. The filter size $(f_w,f_h)$ depends on the convolution layer size,~\eg, the last convolution layer in standard CNNs $\in R^{7\times7}$.

 

This approach imposes no constraints on the network architecture besides having a convolution layer. The input can be a regular image or a pre-extracted convolutional feature. The network output can be logits trained with softmax or a feature embedding trained with a ranking loss. Furthermore, this approach neither changes the original network weights nor requires fine-tuning. Thus, network performance remains intact. The visualization filter is applied only when an attention map is required. Thus, it poses no computational overhead during inference.
L2-CAF visualizes the attention of the last convolutional layer of GoogLeNet within 0.3 seconds.



Section~\ref{sec:approach} describes two variants of the L2-CAF and their mathematical optimization details. The first is the class-oblivious variant illustrated in Figure~\ref{fig:overview}. The second is the class-specific variant for classification networks to localize objects of a specific class. We also present a  technique to reduce the computational cost of the L2-CAF's optimization formulation. We benchmark our approach quantitatively using WSOL for both classification and retrieval architectures.

In summary, the key contributions of this paper are: 

\begin{enumerate}[noitemsep]
\item A novel attention visualization approach for \emph{both} classification and retrieval networks (Sec.~\ref{sec:approach}). This approach achieves state-of-the-art WSOL results using classification architectures (Sec.~\ref{subsec:wsol_cls}). 
\item A modified Grad-CAM to better support WSOL on retrieval networks (Sec. \ref{subsec:wsol_emb}); L2-CAF achieves significant localization improvement margins, up to an absolute 36\%, compared to the vanilla Grad-CAM.
\item A method to visualize attention for video frames that are temporally fused using a recurrent  network (Sec.~\ref{subsec:recurrent_attention}).
\end{enumerate}

	\section{Related Work}

\begin{figure}[t]
	\centering
	\includegraphics[width=0.5\linewidth]{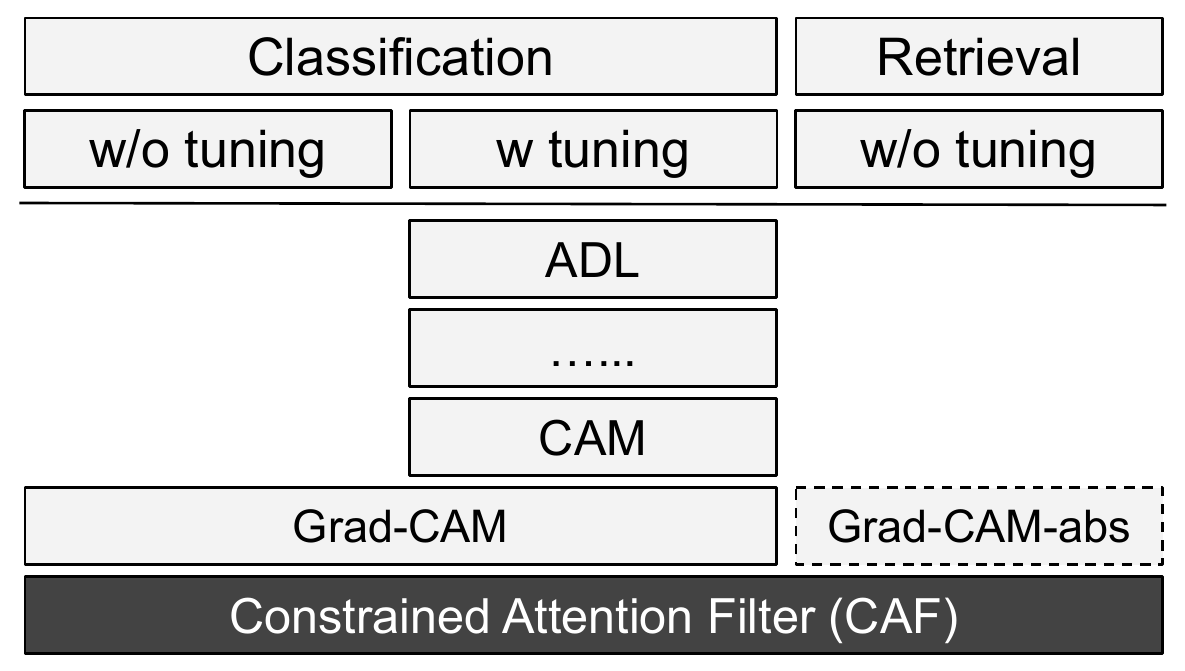}
	\caption{An overview of weakly supervised object localization (WSOL) approaches for classification and retrieval networks. Some approaches impose architectural constraints and require fine-tuning, \eg, CAM and ADL.}
	\label{fig:relatedwork}
\end{figure}

This section briefly reviews weakly supervised object localization (WSOL) for classification networks. Grad-CAM is reviewed in the WSOL retrieval evaluation Section~\ref{subsec:wsol_emb}. The supplementary material reviews different ranking losses (\eg, N-pair). Figure~\ref{fig:relatedwork} presents a high-level categorization of WSOL approaches in terms of (1) supported architectures; (2) whether fine-tuning is required or not? The experiment section provides further one-to-one comparisons.



Classification networks' attention visualization increases interpretability and enables WSOL. CAM~\cite{zhou2016learning} and Grad-CAM~\cite{selvaraju2017grad} identify the most discriminative spatial region. To boost WSOL performance, \cite{singh2017hide,zhang2018adversarial,zhang2018self,choe2019attention} propose architectural modifications to augment the most discriminative region with less-discriminative object regions. This is achieved by fine-tuning a pre-trained network while hiding the most discriminative region stochastically. This forces the network to recognize other informative regions and thus improve WSOL. To detect and hide the most discriminative region while fine-tuning, a network is assumed to use global average pooling (GAP)~\cite{zhou2016learning,singh2017hide,choe2019attention} or an equivalent $1\times 1$ feature reduction convolution layer~\cite{zhang2018adversarial}. This fine-tuning paradigm tends to degrade classification performance.

	\section{Constrained Attention Filter (CAF)}\label{sec:approach}
This section presents two variants for optimizing the proposed L2-CAF. The first variant, \emph{class-oblivious}, works for both classification and retrieval CNNs. It generates a single heatmap per frame. The second variant, \emph{class-specific}, works for classification CNNs and generates class-specific heatmaps per frame. Both variants impose no architectural constraints in terms of spatial pooling (GAP, FCN) or temporal fusing components (RNN, LSTM).

\subsection{Class-Oblivious Variant}\label{subsec:generic_variant}
Given a pre-trained network and an input $x\in R^{W \times H\times 3}$, the last convolution layer provides a feature map $A\in R^{w\times h\times k}$, with size $w\times h$ and $k$ channels.  The network's output $NT(x)$, logits or embedding, depends on discriminative features in $A$. We optimize an L2 normalized filter $f$ to identify the discriminative features to the network's output $NT(x)$.
After multiplying $A$ by the filter $f$ $\left(A \bigodot f = A' \in R^{w\times h\times k}\right)$, the network generates a filtered output $FT(x, f)$. While fixing the network's weights and input, we optimize $f$  to minimize


\begin{equation}\label{eq:class_oblivious}
 L=||NT(x) - FT(x, f)||^2, \quad  \text{subject to}\quad ||f||_2 = 1,
\end{equation}
$FT(x, f)$ equals $ NT(x)$ if and only if $f=f_I=\{1\}^{w\times h}$ which is infeasible due to the unit L2-Norm constraint. 

\noindent\underline{\textbf{Intuition:}} An ideal heatmap can be regarded as a filter that approximates $NT(x)$ by blocking irrelevant features in $A$. Accordingly, we seek a filter $f$ that \textit{spatially} prioritizes convolutional features and \textit{flexibly} captures irregular (\eg, discontinuous) shapes or multiple different agents in a frame. The L2-Norm, a simple \textit{multi-mode} differentiable filter, satisfies these requirements. On account of irrelevant features, the $||f||_2=1$ constraint assigns higher weights to relevant features. Figure~\ref{fig:l2_norm_variants} qualitatively emphasizes the intuition behind the L2-Norm constraint.



%




This formulation (Eq.~\ref{eq:class_oblivious}) is oblivious to the nature of the network's output (logits or embedding), architecture, and input format (RGB image or pre-extracted features). For a given input $x$, the class-oblivious formulation generates a single heatmap. This can be a limitation if the input $x$ contains objects from different classes. The next subsection tackles this limitation by offering an alternative class-specific optimization formulation.

\subsection{Class-Specific Variant}
To support class-specific heatmaps per input, we first assume a classification CNN architecture with class-specific logits. We learn the attention for class $c$ by optimizing the L2-CAF $f$ using the following loss function
\begin{equation}\label{eq:class_specific}
	L_{c}=-FT_c(x, f)+\sum _{ i=0,~\\ i\neq c }^{ N }{ FT_i(x, f) }, \quad  \text{subject to}\quad ||f||_2 = 1,
\end{equation}
where $FT_c(x, f)$ is the filtered output's logit for class $c$ and $N$ is the total number of classes. This loss maximizes the output logit for the intended class $c$ while minimizing the output logits for all other classes.

Figure~\ref{fig:l2_variants} presents a qualitative comparison between the class-oblivious and class-specific variants. For example, the first example shows an image of a butterfly standing on a mastiff's nose. The first image shows the resulting heatmap from optimizing Eq.~\ref{eq:class_oblivious}. The following two images show the result heatmaps from optimizing Eq.~\ref{eq:class_specific} for the mastiff and butterfly classes, respectively. In these examples, the L2-CAF is applied to the last convolutional layer. 


\noindent\underline{\textbf{Technical Details:}} To compute the class-oblivious heatmap for an input $x$, we utilize gradient descent for $l$ iterations. At iteration $i$, $L^{i}$ is computed using the filter $\frac{f^i}{||f^i||_2}$. The filter $f$ is initialized randomly,~\ie, $f^{1} \in [0,1]^{w\times h}$. Gradient descent iteratively updates $f$ to minimize $L$. We terminate when $L$ converges and remains approximately the same for $d$ iterations. Concretely, we terminate the gradient descent at $i=l$ when $\left| L^l - L^{l-d} \right| < \epsilon$ where $\epsilon=10^{-5}$ and $d=50$. This constrained minimization formulation is non-convex, so we also impose a maximum number of iterations $L_{max}$ to avoid oscillating between local minima. After termination, the heatmap is generated by resizing $\frac{|f|^l}{||f^l||_2}$ to the input's size. The same procedure is used for class-specific heatmaps with $L_c$ (Eq.~\ref{eq:class_specific}). For more details, please refer to our released code.

 
\noindent\underline{\textbf{Timing:}} To optimize the small (\eg, ${7\times7}$) L2-CAF using gradient descent, the vanilla L2-CAF requires multiple feed-forward and backpropagation passes through the network. This is affordable for lightweight networks like MobileNet~\cite{howard2017mobilenets} and GoogLeNet (InceptionV1)~\cite{szegedy2015going} but computationally expensive for bulky networks like VGG~\cite{simonyan2014very} and DenseNet~\cite{huang2017densely}. We propose a technique to reduce this cost through (1) making a single feed-forward pass through the whole network to compute the network's output at every layer, (2) optimizing the L2-CAF $f$ using a small subset of layers.

\begin{figure}[t]
	\centering
	\includegraphics[width=0.45\textwidth]{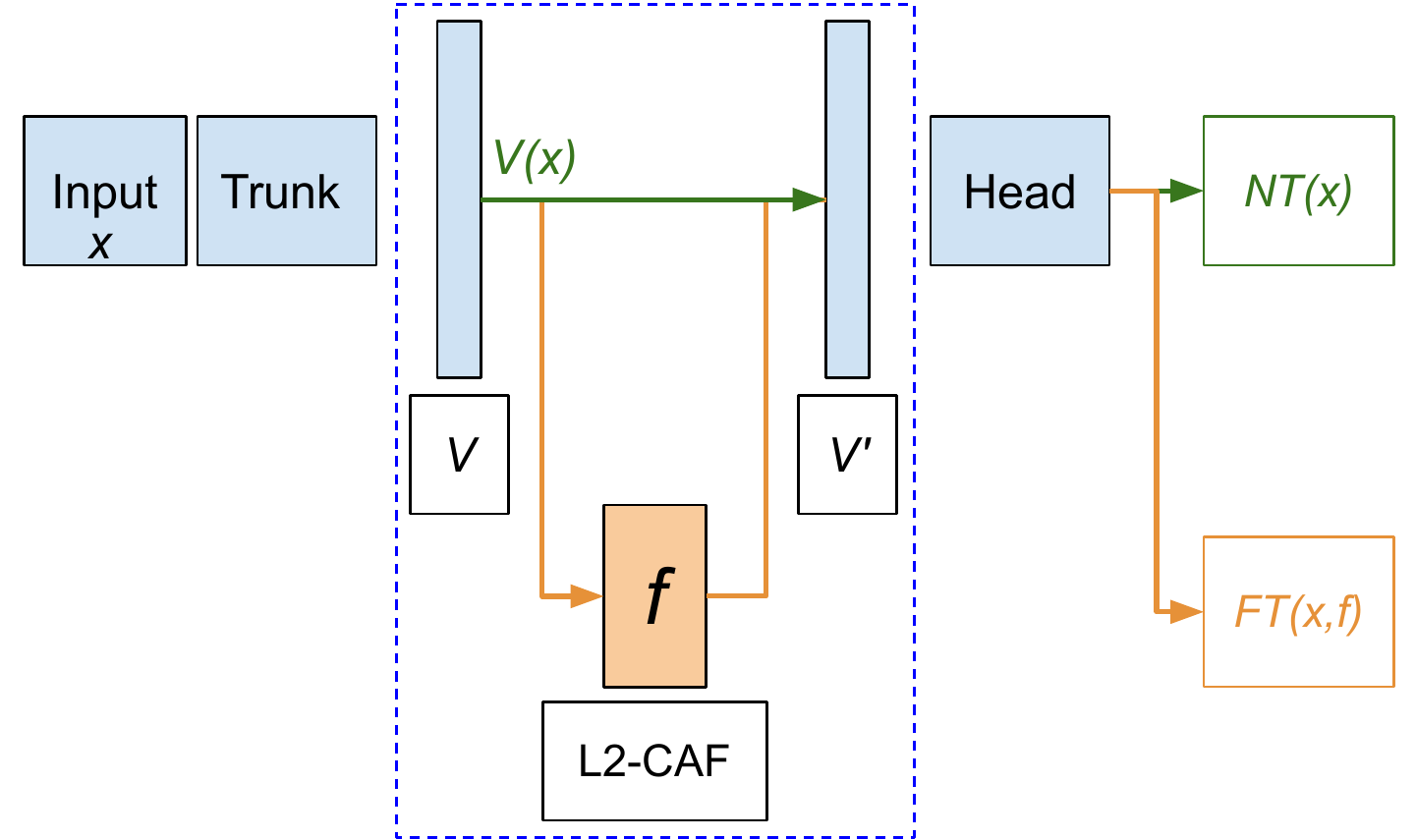} 
	\caption{Reduce the computational optimization cost  of the L2-CAF $f$ by solving an equivalent sub-problem (blue-dashed). Instead of using the network's endpoints ($x$, $NT(x)$), use ($V(x)$, $V'(x)$) to optimize $f$.}
	\label{fig:speed_convergence}
\end{figure}

Figure~\ref{fig:speed_convergence} illustrates this technique. Instead of optimizing the filter $f$ through the network's endpoints ($x$, $NT(x)$), it is equivalent to use the outputs of the direct pre and post layers ($V$, $V'$) to the attention filter. For a given input $x$,  these layers' outputs ($V(x)$, $V'(x)$) require a single feedforward pass through the whole network. Once computed, the loss function from Eq.~\ref{eq:class_oblivious} becomes 
\begin{equation}\label{eq:class_oblivious_fast}
L=||V'(x) - FT(V(x), f)||^2, \quad  \text{subject to}\quad ||f||_2 = 1.
\end{equation}

To generate class-specific heatmaps for a classification network, $V'(x)$ must be the network's logits $NT(x)$. Since it is typical to visualize the attention of the last convolution layer, this formulation skips the overhead of a network's trunk and significantly reduces the computational cost. The speed-up of this technique is quantified through an ablation study. 


The fast L2-CAF approach is a computationally cheaper alternative to sampling \cite{petsiuk2018rise,smilkov2017smoothgrad} and masking~\cite{fong2017interpretable,fong2019understanding} approaches. In addition, the L2-CAF has a smaller set of hyper-parameters. For instance, while both L2-CAF and masking~\cite{fong2017interpretable,fong2019understanding} approaches require a stopping criterion for an optimization problem, Fong~\etal~\cite{fong2019understanding} evaluate multiple mask-sizes per image. Furthermore, the fast L2-CAF works on a small subset of network layers,~\ie, independent of the network backbone. Thus, it compares favorably for video processing. For $3D$ volumes (\eg,  medical images), our optimization problem remains independent of the network size,~\ie, $\in R^{7\times7\times7}$.

\section{Experiments}
The next two subsections present L2-CAF's quantitative evaluation using classification and retrieval networks, respectively. Then, a recurrent retrieval network qualitatively illustrates L2-CAF's potential for video applications. Finally, we present our ablation studies.

\newcommand{\flipud}[1]{\FPeval{\result}{round(100-#1,2)}\result}

\subsection{WSOL Using Classification Networks}\label{subsec:wsol_cls}
The L2-CAF is quantitatively evaluated using WSOL on both \emph{standard} and \emph{fine-tuned} classification architectures. We leverage the ImageNet validation set~\cite{deng2009imagenet} for evaluation on standard architectures. For fined-tuned architectures, we follow ADL~\cite{choe2019attention} evaluation procedure and utilize both ImageNet~\cite{deng2009imagenet} and CUB-200-2011~\cite{wah2011caltech} datasets. In all experiments, we use the fast L2-CAF technique. The loss in Eq.~\ref{eq:class_specific} is minimized using the last convolution layer and the network's logits (before softmax) as endpoints.

\textbf{Evaluation using standard architectures} is performed using both the top-1 and top-5 predictions. Similar to~\cite{selvaraju2017grad}, we obtain the top predictions for every image, then, optimize our filter $f$ to learn the corresponding heatmap for every prediction. Following Zhou~\etal~\cite{zhou2016learning}, we segment the heatmap using a simple thresholding technique. This generates connected segments of pixels; we draw a bounding box around the largest segment. Localization is correct if the predicted class is correct and the intersection over union (IoU) between the ground truth and estimated bounding boxes is $\ge$50\%. Table~\ref{tbl:imgnet_wo_tune_localize_quan} compares L2-CAF and Grad-CAM using three  architectures. Both approaches are applied to the last $7 \times 7$ convolution layer. We fix the architecture and evaluate different localization approaches -- same classification but different localization performance.


\begin{table}[t]
	\centering
	\scriptsize
		\caption{Classification and localization accuracies on the ImageNet (ILSVRC) validation set using standard architectures -- no fine-tuning required.}
	\begin{tabular}{@{}l@{\hspace{7.0\tabcolsep}}c@{\hspace{7.0\tabcolsep}}cc@{}ccc@{}}
		\toprule
		&& \multicolumn{2}{c}{Classification} && \multicolumn{2}{c}{Localization}\\
		\cmidrule{3-4} \cmidrule{6-7}
		Method & Backbone   & Top 1$\uparrow$  & Top 5$\uparrow$ && Top 1$\uparrow$ & Top 5$\uparrow$\\
		\midrule
		Grad-CAM & GoogLeNet~\cite{szegedy2015going}   &  \flipud{28.83}   & \flipud{13.61} && \flipud{55.57} & \flipud{42.500001} \\
		L2-CAF (ours) & GoogLeNet~\cite{szegedy2015going}   & \flipud{28.83}   & \flipud{13.61} &&  \textbf{\flipud{54.52}} & \textbf{\flipud{40.68}} \\
		\midrule
		Grad-CAM & ResNetV2-50~\cite{he2016identity}  &  \flipud{28.49}  & \flipud{13.44}  && \flipud{53.43}  & \flipud{40.04} \\
		L2-CAF (ours) & ResNetV2-50~\cite{he2016identity}   &  \flipud{28.49}  & \flipud{13.44}  && \textbf{\flipud{51.82}}  & \textbf{\flipud{37.62}}  \\
		\midrule
		Grad-CAM & DenseNet-161~\cite{huang2017densely}  & \flipud{21.800001}   & \flipud{8.61} && \flipud{50.72} & \textbf{\flipud{33.43}}  \\
		L2-CAF (ours) & DenseNet-161~\cite{huang2017densely}  &  \flipud{21.800001}  & \flipud{8.61} &&  \textbf{\flipud{50.32}} &  \flipud{34.72} \\
		\bottomrule
	\end{tabular}
	\label{tbl:imgnet_wo_tune_localize_quan}
\end{table}

\noindent\textbf{L2-CAF versus Grad-CAM:} Grad-CAM is $5$ times faster than L2-CAF on DenseNet-161 ($7$ times on GoogLeNet). Both approaches support a large variety of architectures. In terms of localization accuracy, L2-CAF compares favorably to Grad-CAM. Fong and Vedaldi~\cite{fong2017interpretable} explain why gradient-based approaches like Grad-CAM are not optimal for visualization. They show that neural networks' gradients $\frac{\partial y}{\partial x}$ are independent of the input image $x$ for linear classifiers ($y=w x +b; \frac{\partial y}{\partial x}=w$). For non-linear architectures, this problem is reduced but not eliminated. They also show qualitatively that gradient saliency maps contain strong responses in irrelevant image regions. We hypothesize that DenseNet-161's better classification accuracy and, accordingly, better gradient closes the localization performance gap.

 \begin{table}[t]
 	\centering
 	\scriptsize
 	\caption{Classification and localization accuracies on the  CUB-200-2011 test and ImageNet validation split using fine-tuned architectures. The accuracy with an asterisk* indicates that the score is from the original paper.}
 	
 	\begin{tabular}{@{}l@{\hspace{7.0\tabcolsep}}c@{\hspace{7.0\tabcolsep}}c@{\hspace{7.0\tabcolsep}}ccccc@{}}
 		\toprule
 		&  &  & \multicolumn{2}{c}{CUB-200-2011} & & \multicolumn{2}{c}{ImageNet} \\
 		\cmidrule{4-5} 				\cmidrule{7-8}
 		Method & Backbone & Tuning& CLS $\uparrow$ & LOC $\uparrow$ &\phantom{abc}& CLS $\uparrow$ & LOC $\uparrow$\\
 		\midrule
 		CAM & VGG-GAP & GAP & 68.53 & 45.66 && 69.96 & 43.46\\
 		L2-CAF (ours) & VGG-GAP & GAP & 68.53 & 46.01 && 69.96  & 44.09  \\
 		Fuse 2 CAMs & VGG-GAP & ACoL~\cite{zhang2018adversarial} & 71.90 & 45.90* && 67.50 & \textbf{45.83*}\\
 		CAM & VGG-GAP & ADL & 64.16 & 48.27 && 69.58 & 42.93 \\
 		L2-CAF (ours) & VGG-GAP & ADL & 64.16 & \textbf{48.55} && 69.58 & 43.27 \\
 		\midrule
 		CAM & ResNet50-SE & ADL & 78.94 & \textbf{61.71} && 76.218 & 49.90 \\
 		L2-CAF (ours)& ResNet50-SE & ADL & 78.94 & 61.16 && 76.218 & \textbf{50.49} \\
 		\bottomrule
 	\end{tabular}
 	\label{tbl:cub_w_tuning_localize_quan}
 \end{table}




\textbf{Evaluation using fine-tuned architectures} is performed using the top-1 accuracy on fine-tuned architectures (\eg, VGG-GAP~\cite{zhou2016learning}); this follows the evaluation procedure in attention-based dropout layer (ADL)~\cite{choe2019attention}. ADL is the current state-of-the-art method for WSOL. During fine-tuning, ADL applies dropout at multiple network stages. It is not straightforward to determine where to plug these extra dropout layers -- it is network dependent. Therefore, we leverage their publicly released VGG-GAP and ResNet50-SE implementations to evaluate our approach. The ACoL performance is reported from the original paper.

Table~\ref{tbl:cub_w_tuning_localize_quan} presents a quantitative evaluation using CUB-200-2011 and ImageNet datasets. The first column denotes the object localization approach,~\eg, CAM versus L2-CAF. Grad-CAM is dropped because it is equivalent to CAM when a GAP layer is utilized~\cite{selvaraju2017grad}. In the second column (backbone), all the architectures utilize a global average pooling or an equivalent surrogate~\cite{zhang2018adversarial}. The third column denotes the fine-tuning approaches considered: GAP~\cite{zhou2016learning}, ACol~\cite{zhang2018adversarial}, ADL~\cite{choe2019attention}. We fine-tune the VGG-GAP architecture with both GAP and ADL. L2-CAF consistently outperforms CAM's localization on the ImageNet validation set. 

\noindent\textbf{Relation with WSOL approaches (\eg, ADL):} To generate class activation maps (CAMs), WSOL approaches employ a global average pooling layer (GAP)~\cite{zhou2016learning,singh2017hide,zhang2018self,choe2019attention}, or equivalent~\cite{zhang2018adversarial}. L2-CAF relaxes this architectural requirement. Thus, while supporting previous WSOL approaches, L2-CAF introduces a new degree of freedom. This flexibility is vital to explore attention visualization and WSOL beyond standard supervised classification networks.

	\subsection{WSOL Using Retrieval Networks}\label{subsec:wsol_emb}

Weakly supervised object localization provides a quantitative evaluation metric for attention visualization approaches. The ability to localize attention for various architectures is a core advantage of L2-CAF. In this subsection, we quantitatively evaluate L2-CAF against Grad-CAM. We employ the class oblivious formulation (Eq.~\ref{eq:class_oblivious})  using the last convolution layer and the raw embedding (before unit-circle normalized) as endpoints.

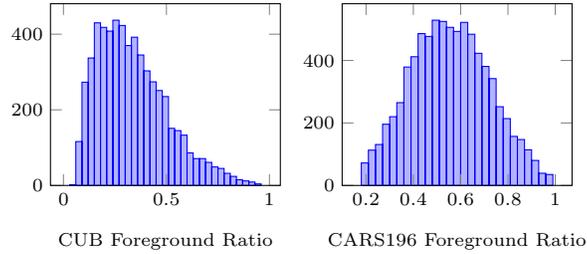
\begin{figure}[t]
	\centering
	\scriptsize
	\begin{tikzpicture}
\begin{axis}[
ymin=0,
height=4.0cm,
xlabel={CUB Foreground Ratio},
area style,
]
\addplot+[ybar interval,mark=no] plot coordinates { 
	(0.030000,2)
	(0.060000,116)
	(0.090000,273)
	(0.120000,337)
	(0.150000,430)
	(0.180000,418)
	(0.210000,408)
	(0.240000,437)
	(0.270000,423)
	(0.300000,370)
	(0.330000,392)
	(0.360000,345)
	(0.390000,303)
	(0.420000,274)
	(0.450000,251)
	(0.480000,235)
	(0.510000,151)
	(0.540000,145)
	(0.570000,133)
	(0.600000,86)
	(0.630000,71)
	(0.660000,71)
	(0.690000,61)
	(0.720000,49)
	(0.750000,45)
	(0.780000,32)
	(0.810000,24)
	(0.840000,15)
	(0.870000,12)
	(0.900000,9)
	(0.930000,4)
	(0.960000,2)
};
\end{axis}
\end{tikzpicture}~
\begin{tikzpicture}
\begin{axis}[
ymin=0,
height=4.0cm,
	xlabel={CARS196 Foreground Ratio},
area style,
]
\addplot+[ybar interval,mark=no] plot coordinates { 
(0.180000,72)
(0.210000,113)
(0.240000,131)
(0.270000,196)
(0.300000,220)
(0.330000,265)
(0.360000,379)
(0.390000,412)
(0.420000,486)
(0.450000,477)
(0.480000,529)
(0.510000,525)
(0.540000,506)
(0.570000,493)
(0.600000,522)
(0.630000,484)
(0.660000,423)
(0.690000,381)
(0.720000,342)
(0.750000,252)
(0.780000,214)
(0.810000,157)
(0.840000,147)
(0.870000,114)
(0.900000,80)
(0.930000,39)
(0.960000,35)
(0.990000,7)
 };
\end{axis}
\end{tikzpicture}
	\caption{Histogram of the foreground objects' bounding box size relative to the whole image in CUB  and CARS196 datasets. CUB birds tend to occupy less than 50\% of the whole image (left-skewed), while the Stanford cars are normally distributed.}
	\label{fig:histogram}
\end{figure}

\noindent\textbf{Datasets:} We employ CUB-200-2011 birds~\cite{wah2011caltech} and Stanford CARS196~\cite{krause20133d} retrieval datasets,~\ie, standard retrieval datasets~\cite{oh2016deep,wang2017deep,chen2017beyond,wu2017sampling}. Both datasets provide the ground truth bounding box annotation. They pose several challenges for foreground objects' localization. Birds are not naturally rectangular; discriminative parts (\eg, head~\cite{chen2018looks})  occupy a small part of the body. Cars pose a similar challenge in terms of relatively smaller discriminative parts (\eg, wheel) relative to the whole body. Figure~\ref{fig:histogram} depicts the ratio of the ground truth bounding box to the whole image size for both datasets.


\noindent\textbf{Evaluation metrics:} \underline{For retrieval}, we utilize both Recall@1 (R@1) and the Normalized Mutual Information (NMI) metrics. \underline{For localization}, we follow the same evaluation procedure in \cite{zhou2016learning,selvaraju2017grad} for classification networks. We replace the top-1 by R@1 metric to decide if the network's output is correct or not. The same IoU $> 50\%$ criterion is used to evaluate localization.

\noindent\textbf{Vanilla Grad-CAM baseline:} To evaluate L2-CAF quantitatively, we extend the classification Grad-CAM to deal with retrieval networks. The Grad-CAM  class-discriminative localization map $M^c$ has been proposed as follows
\begin{align}\label{eq:grad_cam}
M^c = \textit{RELU}\left(\sum_k{\alpha^c_k A^k}\right)\\
\alpha^c_k =\frac{1}{w\times h} \sum_{i=0}^w \sum_{j=0}^h \frac{\partial y^c}{\partial A^k_{i,j}},
\end{align}
where $M^c \in R^{w\times h}$ for any class $c$, $A \in R^{w\times h\times k}$ is a convolutional feature map with $k$ channels. $\alpha^c \in R^k$ quantifies the $k^{th}$ channel's importance for a target class $c$. Basically, $\sum_k{\alpha^c_k A^k}$ provides a weighted sum of the feature maps ($A$) for class $c$.
 $\alpha^c_k$ is computed using the gradient of the score for class $y^c$ with respect to the feature maps $A^k$.

To support a retrieval network, we utilize the gradient of the output embedding $\frac{\partial y}{\partial A^k_{i,j}}$ instead of the class score $\frac{\partial y^c}{\partial A^k_{i,j}}$ as follows 
\begin{equation}
\alpha^y_k =\frac{1}{w\times h} \sum_{i=0}^w \sum_{j=0}^h \frac{\partial y}{\partial A^k_{i,j}},
\end{equation}
we denote this formulation as \emph{vanilla Grad-CAM} for retrieval. We compute $\frac{\partial y}{\partial A^k_{i,j}}$ using tf.gradients~\cite{tf2020grad}.

\newcommand{\upmargin}[1]{\FPeval{\result}{round(#1,2)}+\result}

\begin{table}[t]
	\centering
	\scriptsize
		\caption{Triplet (TL) and N-pair (NP) losses' quantitative retrieval evaluation using NMI and Recall@1 on CUB-200-2011 and CARS196. Quantitative localization accuracy evaluation using the 0.5 intersection over union (IoU) criterion. $\triangle$ column indicates the absolute localization improvement margin relative to the vanilla Grad-CAM.}
		
	
	\begin{tabular}{@{}l@{\hspace{7.0\tabcolsep}}c@{\hspace{7.0\tabcolsep}}c @{\hspace{7.0\tabcolsep}} cc c c@{\hspace{5.0\tabcolsep}}c c cc c c@{\hspace{5.0\tabcolsep}}c@{}}
		\toprule
		& & & \multicolumn{5}{c}{CUB-200-2011} &\phantom{} & \multicolumn{5}{c}{CARS196} \\
		\cmidrule{4-8}  		\cmidrule{10-14}
		& & & \multicolumn{2}{c}{Retrieval} && \multicolumn{2}{c}{Localization} & & \multicolumn{2}{c}{Retrieval} && \multicolumn{2}{c}{Localization}\\
		\cmidrule{4-5} \cmidrule{7-8} \cmidrule{10-11} \cmidrule{13-14}
		Method & Backbone & Loss  & NMI$\uparrow$ & R@1$\uparrow$  && LOC$\uparrow$ & $\triangle$ 
													 && NMI$\uparrow$ & R@1$\uparrow$  && LOC$\uparrow$ & $\triangle$  \\
		\midrule
		Grad-CAM& GoogLeNet & TL & 0.582 & 47.75 &  & \flipud{79.372046}& - && 0.532 & 54.55 &  & \flipud{67.617759}& -\\
		
		Grad-CAM-abs& GoogLeNet & TL & 0.582 & 47.75 &  & \flipud{77.042539} & \upmargin{79.372046-77.042539} & & 0.532 & 54.55 &  & \flipud{53.794121}& \upmargin{67.617759-53.794121} \\
		L2-CAF (ours)& GoogLeNet & TL & 0.582 & 47.75 &  & \textbf{\flipud{70.374747}}& \upmargin{79.372046-70.374747} & & 0.532 & 54.55 &  & \textbf{\flipud{45.898413}}& \upmargin{67.617759-45.898413}\\ \midrule
		
		Grad-CAM& ResNet & TL & 0.601 & 50.06 & & \flipud{83.845375} & - & & 0.565 & 61.55 & & \flipud{68.23269} & -\\
		Grad-CAM-abs& ResNet & TL & 0.601 & 50.06 & & \flipud{70.509791} & \upmargin{83.845375-70.509791} & & 0.565 & 61.55 & & \flipud{43.684664} & \upmargin{68.23269-43.684664}\\
		L2-CAF (ours)& ResNet & TL & 0.601 & 50.06 & & \textbf{\flipud{60.719109}} & \upmargin{83.845375-60.719109} && 0.565 & 61.55 & & \textbf{\flipud{38.728324}} & \upmargin{68.23269-38.728324}\\ \midrule
		
		Grad-CAM& GoogLeNet & NP  & 0.583  & 48.95 & & \flipud{85.871033} & - & & 0.597  & 65.23 & & \flipud{71.713196} & -\\
		Grad-CAM-abs& GoogLeNet & NP  & 0.583  & 48.95 & & \flipud{80.131668} & \upmargin{85.871033-80.131668} & & 0.597  & 65.23 & & \flipud{44.988316} & \upmargin{71.713196-44.988316}\\
		L2-CAF (ours) & GoogLeNet & NP  & 0.583  & 48.95 & & \textbf{\flipud{69.496962}} & \upmargin{85.871033-69.496962} & & 0.593  & 65.23 & & \textbf{\flipud{35.149428}} & \upmargin{71.713196-35.149428}\\ \midrule
		
		Grad-CAM& ResNet & NP & 0.580 & 47.92 && \flipud{88.082377} & - & & 0.609 & 67.61 && \flipud{67.580863} & - \\
		Grad-CAM-abs& ResNet & NP & 0.580 & 47.92 && \flipud{73.328832} & \upmargin{88.082377-73.328832} & & 0.609 & 67.61 && \flipud{38.383963} & \upmargin{67.580863-38.383963}\\
		L2-CAF (ours) & ResNet & NP & 0.580 & 47.92 && \textbf{\flipud{61.309926}} & \upmargin{88.082377-61.309926} & & 0.609 & 67.61 && \textbf{\flipud{32.65281}} & \upmargin{67.580863-32.65281}\\
		\bottomrule
	\end{tabular}
	\label{tbl:localize__cub_quan}
\end{table}

\noindent\textbf{Grad-CAM-abs baseline:} The Vanilla Grad-CAM is largely inferior for retrieval networks because of the \textit{RELU} in Eq.~\ref{eq:grad_cam}. \textit{RELU} is introduced for classification networks to emphasize feature maps that have a positive influence on the class of interest $y^c$, assuming pixels with negative gradient belong to other classes. This assumption is valid for classification but invalid for retrieval. Therefore, we further modify the Grad-CAM formulation by replacing the $RELU$ with the absolute function $abs$. This Grad-CAM-abs baseline is defined as follows 
\begin{equation}\label{eq:grad_cam_abs}
M_{abs}^{y} = \textit{abs}\left(\sum_k{ \alpha^y_k A^k}\right).
\end{equation}

\noindent\textbf{Implementation details} are reported in the supplementary material.

%


\newcommand{\GradNormImgSize}{0.15}
\begin{figure}[t]
	\centering
	\scriptsize
	\setlength\tabcolsep{2.0pt} 
	\renewcommand{\arraystretch}{1.0}	

	\begin{tabular}{@{}cccc@{}}
		Image & Grad-CAM & Grad-CAM-abs & L2-CAF \\
		\includegraphics[width=\GradNormImgSize\textwidth,height=\GradNormImgSize\textwidth]{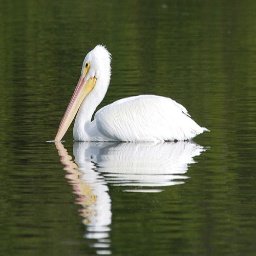} &
		\includegraphics[width=\GradNormImgSize\textwidth,height=\GradNormImgSize\textwidth]{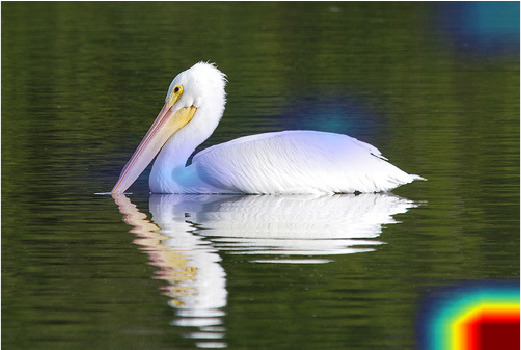} & \includegraphics[width=\GradNormImgSize\textwidth,height=\GradNormImgSize\textwidth]{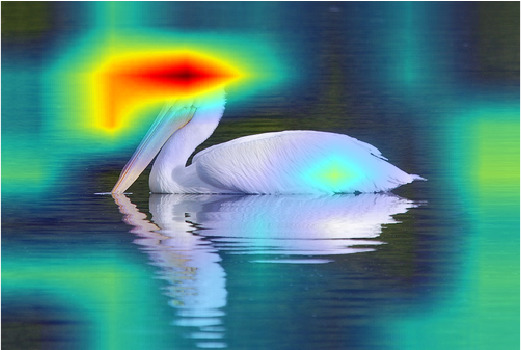} &
		\includegraphics[width=\GradNormImgSize\textwidth,height=\GradNormImgSize\textwidth]{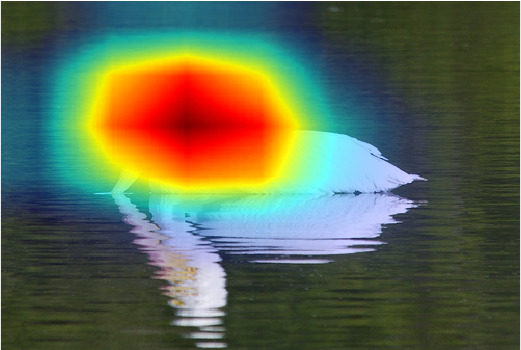} \\
		
		\includegraphics[width=\GradNormImgSize\textwidth,height=\GradNormImgSize\textwidth]{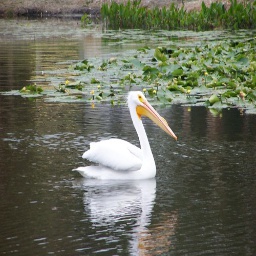} &
		\includegraphics[width=\GradNormImgSize\textwidth,height=\GradNormImgSize\textwidth]{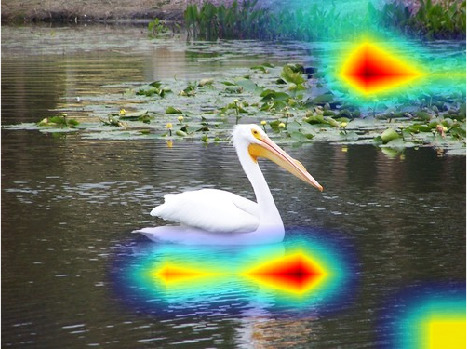} & \includegraphics[width=\GradNormImgSize\textwidth,height=\GradNormImgSize\textwidth]{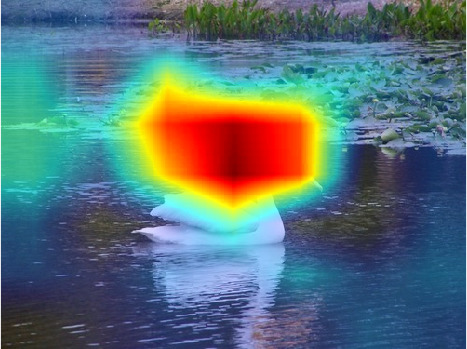} &
		\includegraphics[width=\GradNormImgSize\textwidth,height=\GradNormImgSize\textwidth]{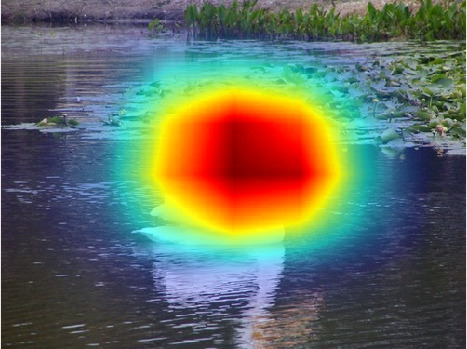} \\

	\end{tabular}
	\caption{Qualitative attention evaluation for different visualization approaches on retrieval networks. Both Grad-CAM variants suffer near images' corners.}
	\label{fig:retrieval_qual}
\end{figure}

\noindent\textbf{Results:} Table~\ref{tbl:localize__cub_quan} presents a quantitative evaluation for both retrieval and localization performance. ResNet-50 has more parameters than GoogLeNet; and is marginally better in terms of retrieval. Generally, N-pair loss outperforms triplet loss. Cars are rectangular and thus simpler than CUB birds for bounding box localization. The localization error is highly correlated and upper-bounded by the retrieval performance (R@1). Grad-CAM-abs outperforms the vanilla Grad-CAM for retrieval. L2-CAF brings further localization improvement.

 Figure~\ref{fig:retrieval_qual} qualitatively compares different localization approaches. We found that feature maps at the images' corners can have a high positive gradient, while the feature maps at the foreground object can have a high negative gradient. It is a common practice to embed images into the unit-circle,~\ie, some images are embedded in the negative space. When this happens, the vanilla Grad-CAM ignores the foreground objects. Grad-CAM-abs handles negative gradient better but still suffers around the corners. Grad-CAM inferior behavior around the corner is qualitatively reported in~\cite{fong2019understanding}. This undesirable behavior degrades Grad-CAM's WSOL performance.
 Figure~\ref{fig:qual_emb_eval} shows a qualitative localization evaluation on both datasets.  For CUB-200-2011, our estimated bounding box (blue) tends to be centered around the birds' heads.




\newcommand{\embeddingImgSize}{0.09}
\begin{figure}[t]
	\setlength\tabcolsep{1.0pt} 
	\renewcommand{\arraystretch}{1.0}	
	
	\centering
	\includegraphics[width=\embeddingImgSize\textwidth,height=\embeddingImgSize\textwidth]{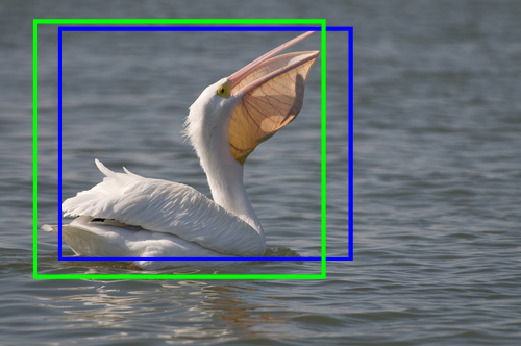} 
	\includegraphics[width=\embeddingImgSize\textwidth,height=\embeddingImgSize\textwidth]{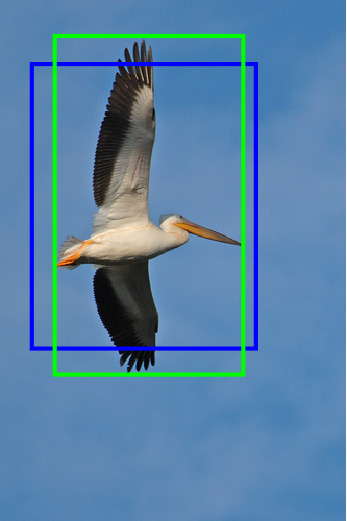}
	\includegraphics[width=\embeddingImgSize\textwidth,height=\embeddingImgSize\textwidth]{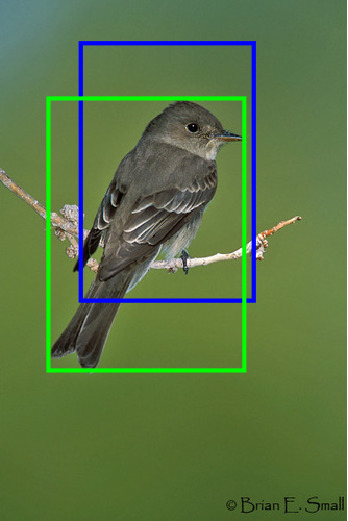} 
	\includegraphics[width=\embeddingImgSize\textwidth,height=\embeddingImgSize\textwidth]{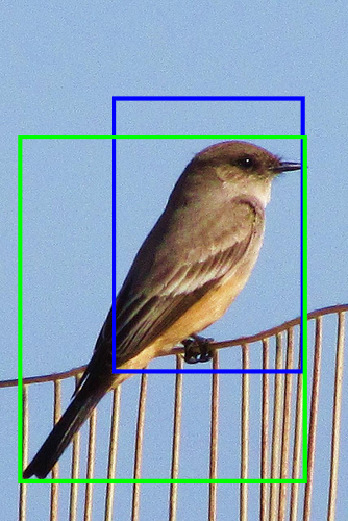}
	\includegraphics[width=\embeddingImgSize\textwidth,height=\embeddingImgSize\textwidth]{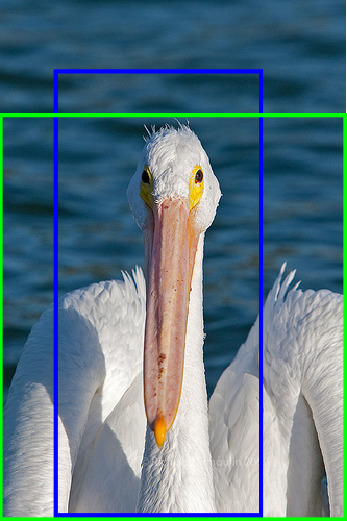} \hfill
	\includegraphics[width=\embeddingImgSize\textwidth,height=\embeddingImgSize\textwidth]{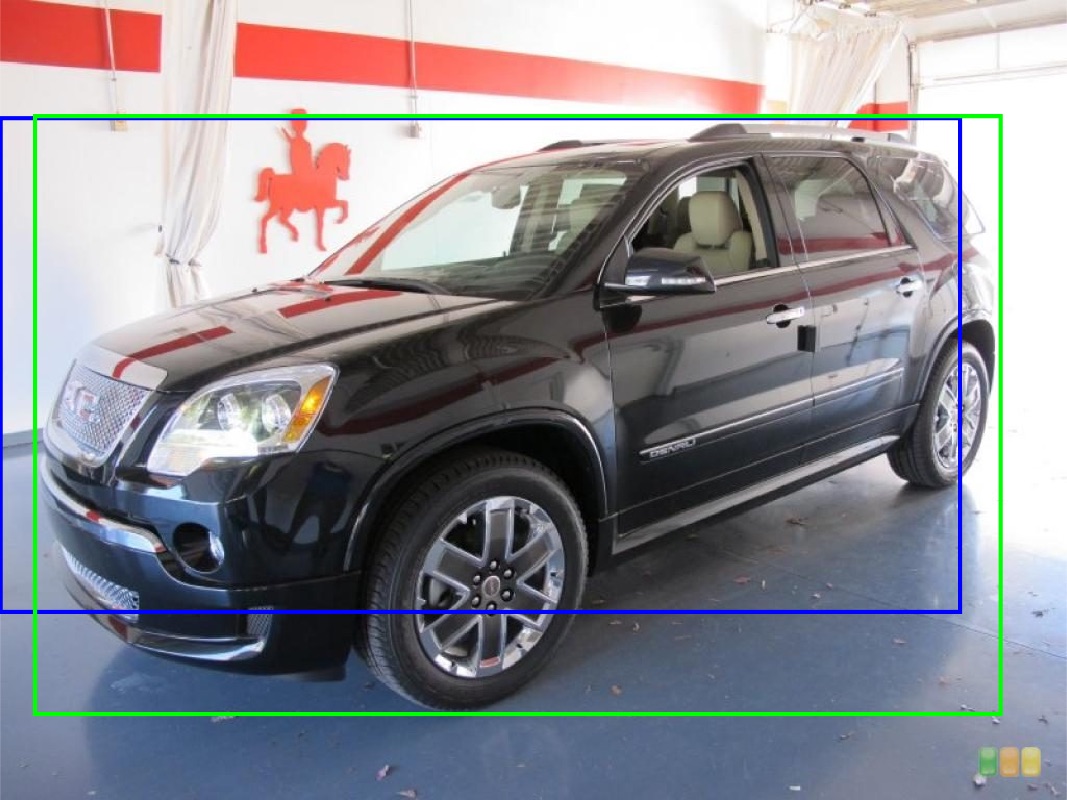} 
	\includegraphics[width=\embeddingImgSize\textwidth,height=\embeddingImgSize\textwidth]{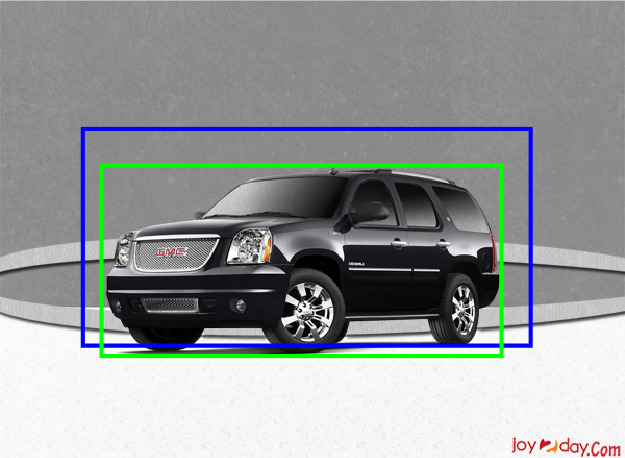}
	\includegraphics[width=\embeddingImgSize\textwidth,height=\embeddingImgSize\textwidth]{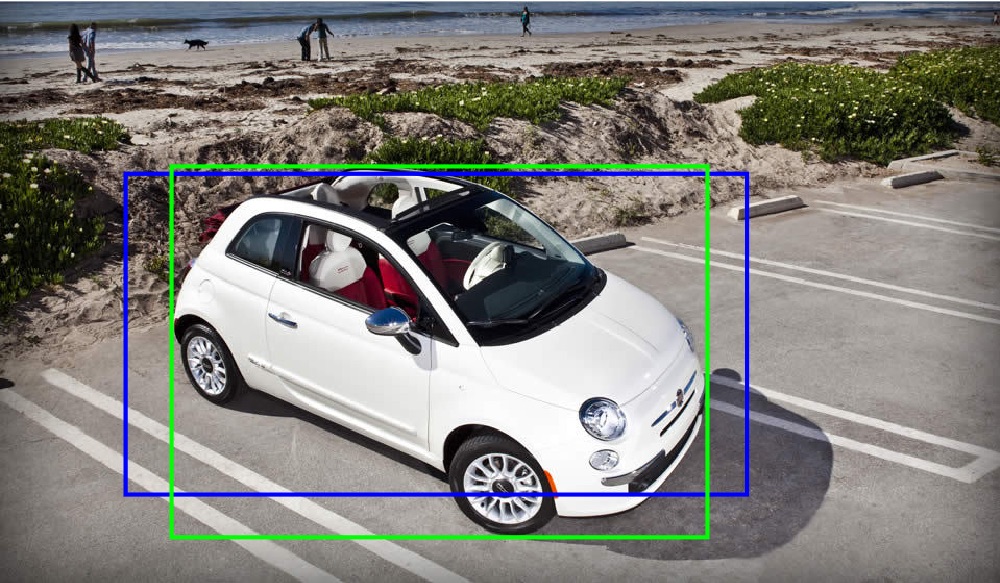} 
	\includegraphics[width=\embeddingImgSize\textwidth,height=\embeddingImgSize\textwidth]{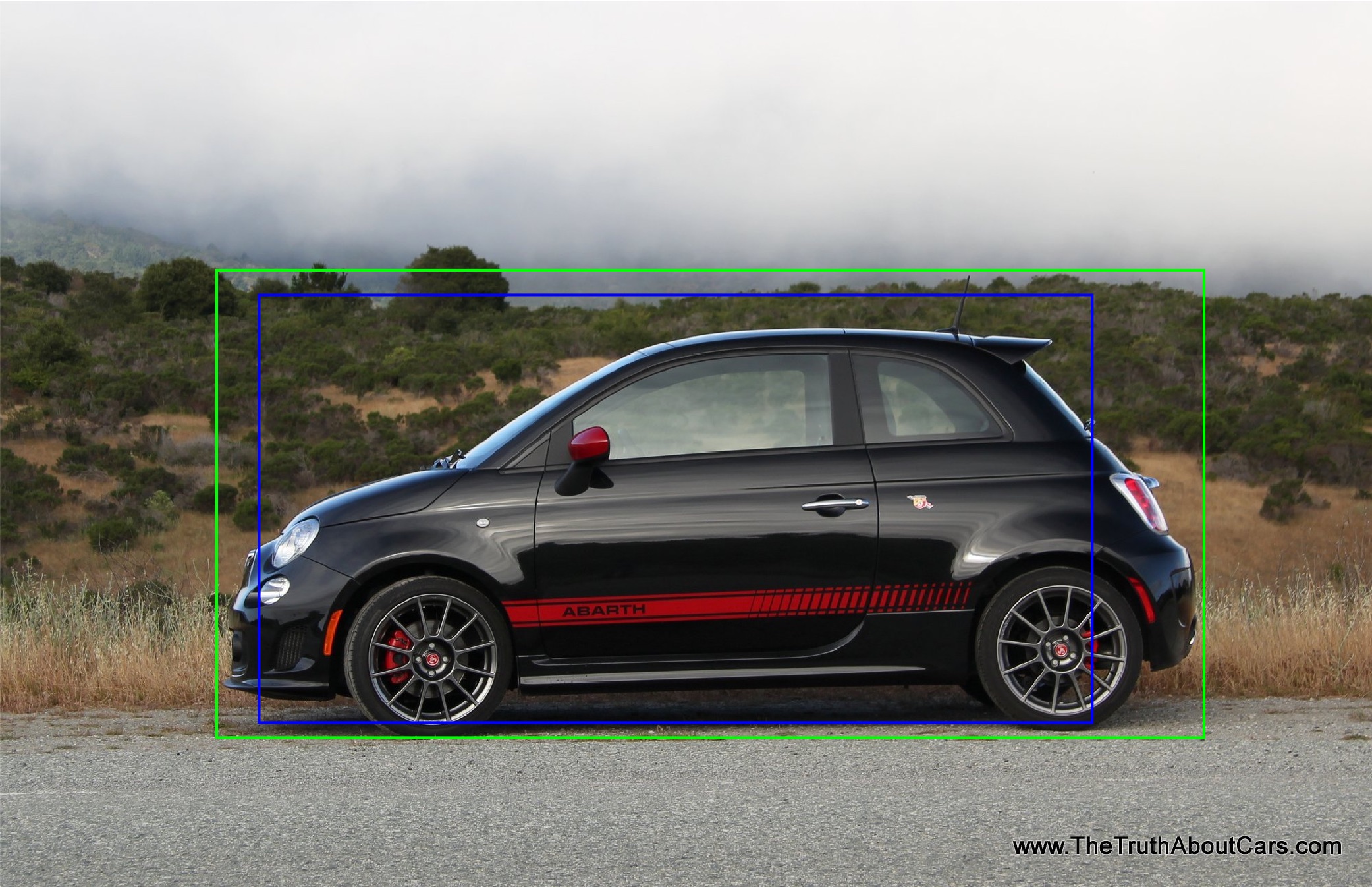}
	\includegraphics[width=\embeddingImgSize\textwidth,height=\embeddingImgSize\textwidth]{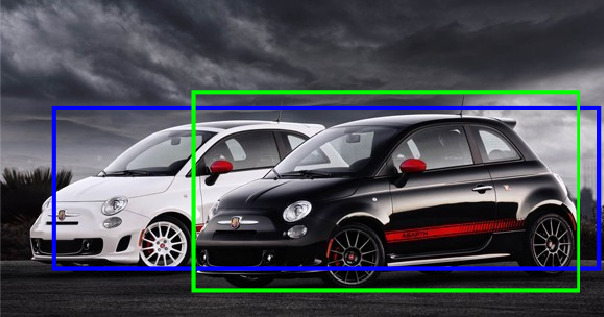} 
	\caption{Qualitative localization evaluation on CUB-200-2011 and  CARS196 using a retrieval network trained with  a triplet loss. The green and blue bounding boxes indicate the ground truth and the L2-CAF bounding boxes, respectively.}
	\label{fig:qual_emb_eval}
\end{figure}

	\subsection{Recurrent Networks' Attention}\label{subsec:recurrent_attention}

\begin{figure}[t]
	\centering
	\scriptsize
	\includegraphics[width=0.40\textwidth]{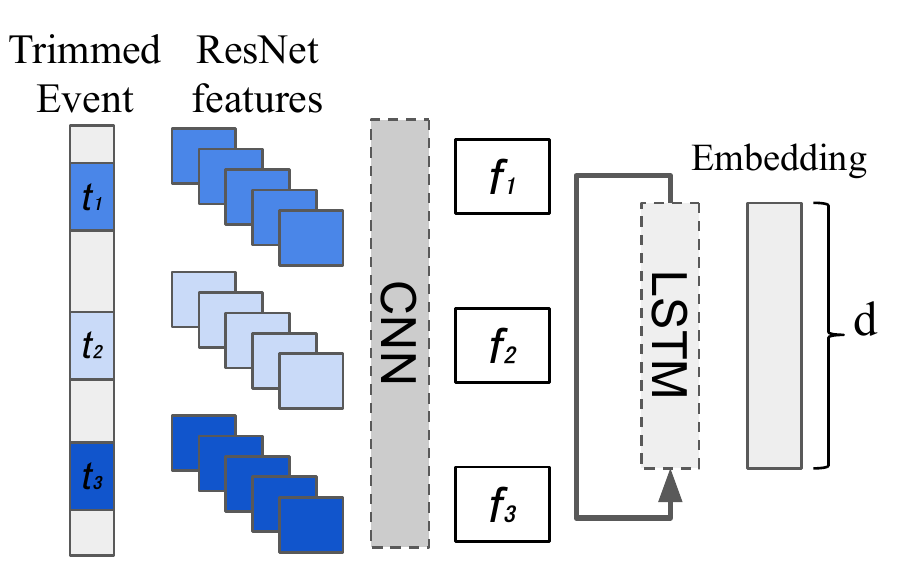}
	\caption{A convolution architecture to embed autonomous navigation videos. This network employs a ranking loss to learn a feature embedding and a recurrent layer for temporal modeling. The CNN layer is shared across the three frames. The attention filters ($f_1,f_2,f_3$) are used during attention visualization only.}
	\label{fig:honda_arch}
\end{figure}

This subsection illustrates how to visualize attention for temporally fused video frames through the Honda driving dataset (HDD)~\cite{ramanishka2018toward}. HDD is a video dataset for reasoning about drivers' actions (events) like crossing intersections, making left and right turns. A key objective is modeling the subtle intra-action (events) variations without explicit fine-grained labeling. For instance, an autonomous navigation application with a left-turn query video should differentiate smooth left-turns maneuvers from those interrupted by crossing pedestrians. A retrieval network models these intra-action variations through a feature embedding.





Figure~\ref{fig:honda_arch} presents a recurrent retrieval network for video embedding. Given a trimmed video event, three frames are sampled at $t_1$, $t_2$, and $t_3$. To enable a large training mini-batch for triplet loss, a pre-trained ResNet is employed to extract convolutional features for every frame. The extracted ResNet features are fed into a trainable shallow CNN.  The resulting convolutional features are temporally fused using an LSTM~\cite{funahashi1993approximation,hochreiter1997long}. 



After training, we employ three L2-CAF filters ($f_1$, $f_2$, and $f_3$) to visualize attention,~\ie, one filter per frame. These filters are inserted between the shallow CNN and the LSTM layers \emph{during inference only}. To ground attention in each frame, we optimize each filter independently. Concretely, we pass the first frame's features through $f_1$ and optimize $f_1$ while feeding the second and third frames' features normally, \ie, $f_2$ and $f_3$ are inactive. After $f_1$ converges, we deactivate it and optimize the next filter $f_2$ and so on. After optimizing all filters, each filter provides an attention map for the corresponding frame.

\newcommand{\autoImgSize}{0.13}
\begin{figure}[t]
	\centering
	\scriptsize
	\setlength\tabcolsep{1.0pt} 
	\renewcommand{\arraystretch}{1.0}	
	
	\begin{tabular}{@{}ccc@{}}
		$t_1$ & $t_2$ & $t_3$ \\
		\includegraphics[width=\autoImgSize\textwidth,height=\autoImgSize\textwidth]{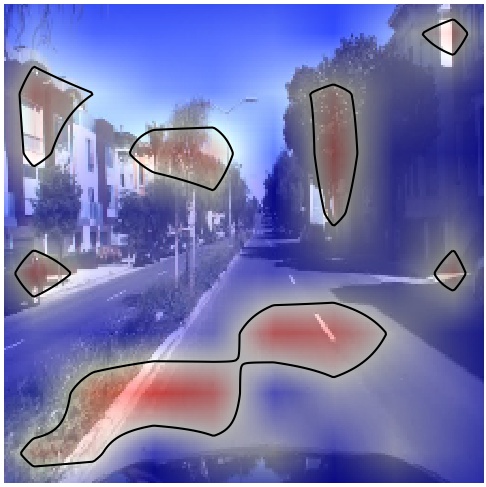} &
		\includegraphics[width=\autoImgSize\textwidth,height=\autoImgSize\textwidth]{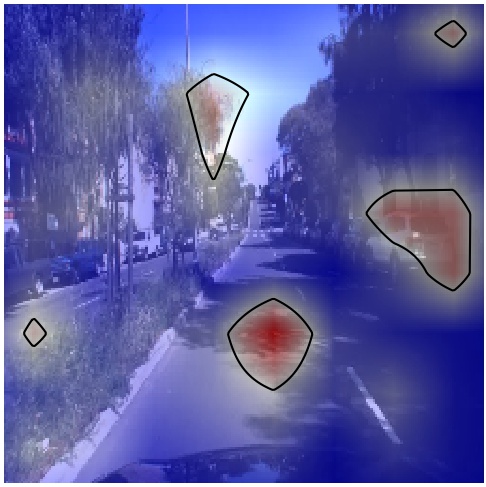} & \includegraphics[width=\autoImgSize\textwidth,height=\autoImgSize\textwidth]{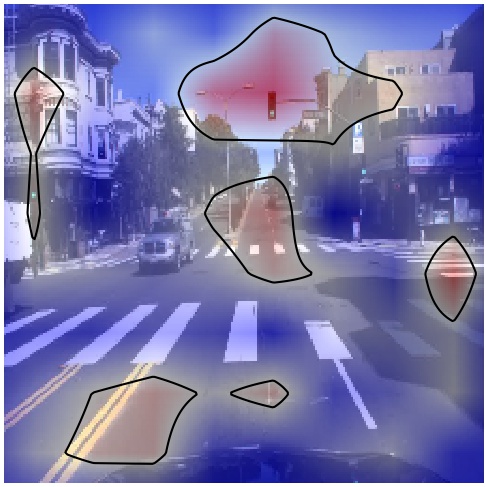} \\
		
		\includegraphics[width=\autoImgSize\textwidth,height=\autoImgSize\textwidth]{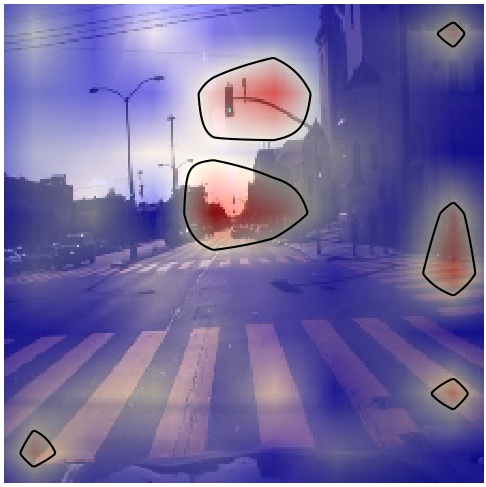} &
		\includegraphics[width=\autoImgSize\textwidth,height=\autoImgSize\textwidth]{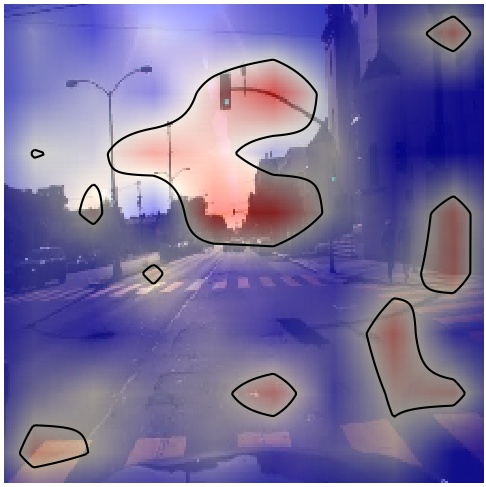} & \includegraphics[width=\autoImgSize\textwidth,height=\autoImgSize\textwidth]{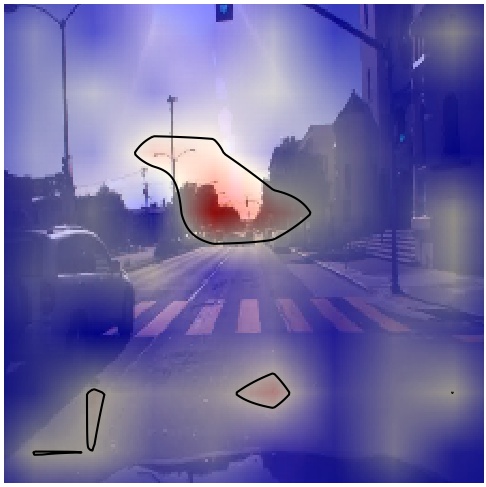} \\
		
		\includegraphics[width=\autoImgSize\textwidth,height=\autoImgSize\textwidth]{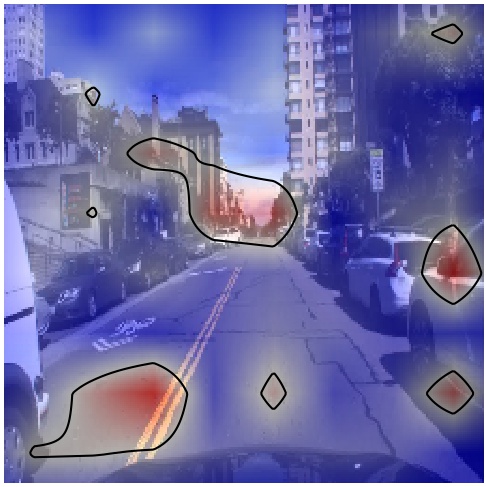} &
		\includegraphics[width=\autoImgSize\textwidth,height=\autoImgSize\textwidth]{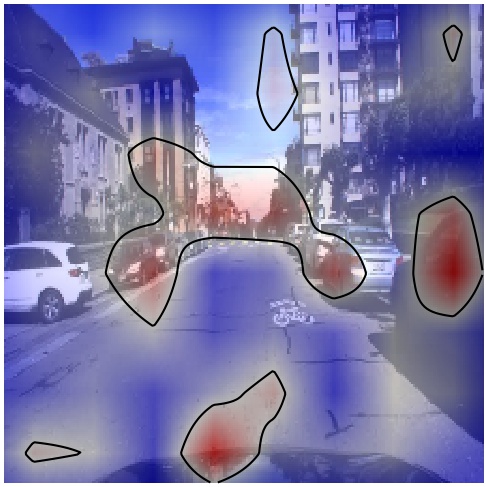} & \includegraphics[width=\autoImgSize\textwidth,height=\autoImgSize\textwidth]{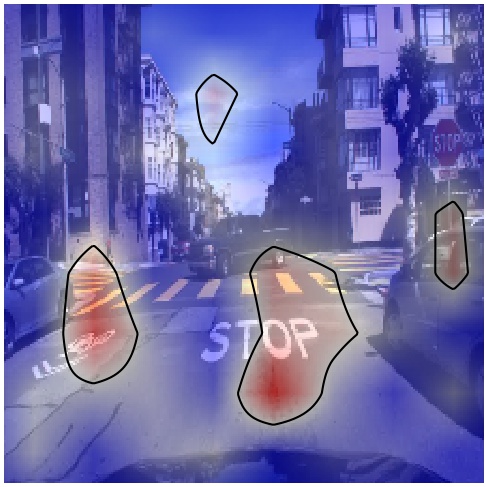} \\
	\end{tabular}
	\caption{The recurrent network's attention visualization at different time steps using heatmaps. Each row depicts three frames sampled from an action video. Contours highlight regions with higher attention. The network attends to the spatial locations of traffic lights and road signs. This figure is best viewed on a screen (color and zoom).}
	\label{fig:auto_qual}
\end{figure}

Figure~\ref{fig:auto_qual} presents our qualitative evaluation. In the first row at $t_3$, the network attention is drawn to the traffic lights and double yellow lane marks. Similarly, the second row shows attention drawn toward the traffic light at $t_{1,2}$. The final row shows an interesting case at $t_3$ where the attention is drawn to the stop sign and also to the frame's top center, which is the typical location for a traffic light. Through visualization, we can see that the network uses traffic lights, signs, and road signs as discriminative features. We found that a Mast Arm (L-shaped) traffic light is easier to detect by a neural network compared to a straight pole traffic light. The variable height of a straight pole traffic light poses a challenge for neural networks. For instance, the network attends to the right side of the frame multiple times at different heights in the second row at $t_{1,2}$.  




	
	\subsection{Ablation Study}\label{subsec:ablation}
This subsection provides sanity checks for saliency maps~\cite{adebayo2018sanity}, then evaluates alternative attention constraints, and finally presents a timing analysis.

\newcommand{\SmallVaraintsImgSize}{0.10}
\begin{figure}[t]
	\centering
	\tiny
	\setlength\tabcolsep{1.0pt} 
	\renewcommand{\arraystretch}{1.0}	
	\begin{tabular}{@{}c c cc c cc@{}}
		Pretrained &\phantom{a}& \multicolumn{2}{c}{Different Random Logits}  &\phantom{a}& \multicolumn{2}{c}{Different Random Weights}\\
		
		\includegraphics[width=\SmallVaraintsImgSize\textwidth,height=\SmallVaraintsImgSize\textwidth]{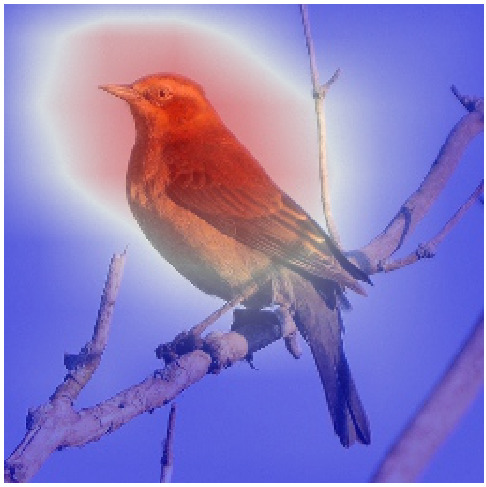} &&
		\includegraphics[width=\SmallVaraintsImgSize\textwidth,height=\SmallVaraintsImgSize\textwidth]{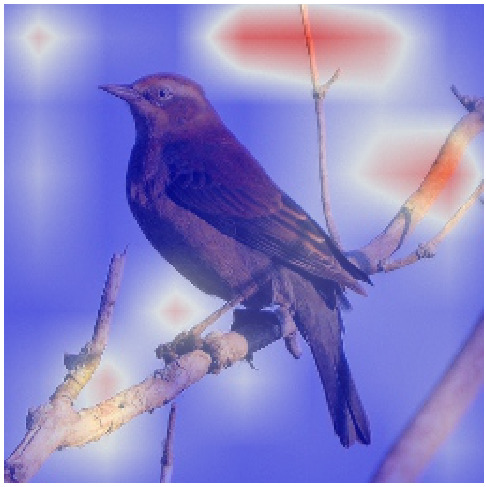} & \includegraphics[width=\SmallVaraintsImgSize\textwidth,height=\SmallVaraintsImgSize\textwidth]{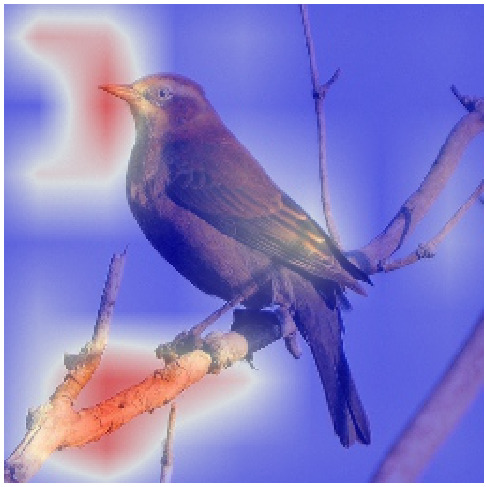} && \includegraphics[width=\SmallVaraintsImgSize\textwidth,height=\SmallVaraintsImgSize\textwidth]{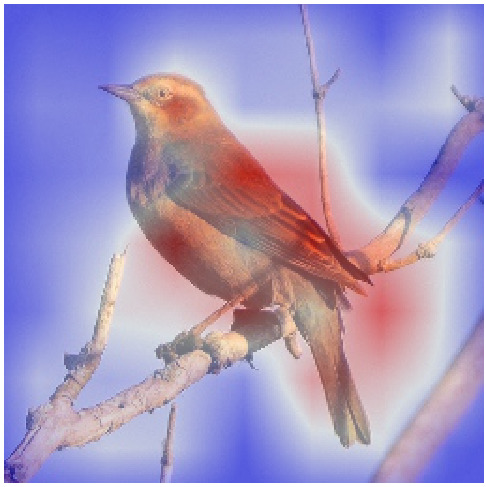} & \includegraphics[width=\SmallVaraintsImgSize\textwidth,height=\SmallVaraintsImgSize\textwidth]{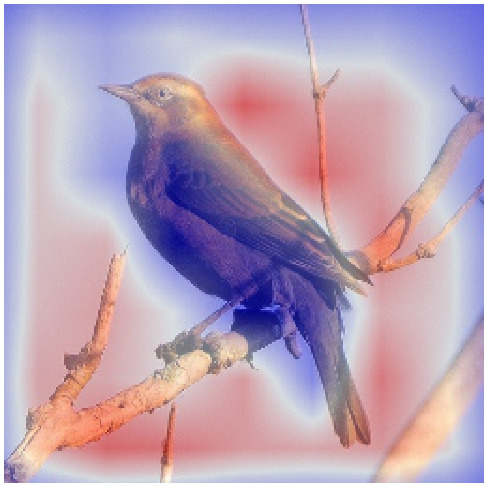} \\
		
		\includegraphics[width=\SmallVaraintsImgSize\textwidth,height=\SmallVaraintsImgSize\textwidth]{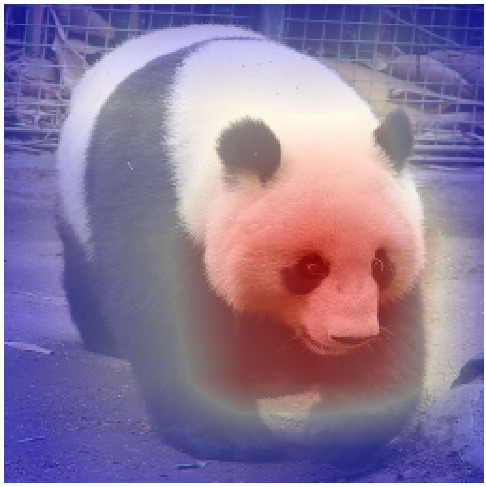} &&
		\includegraphics[width=\SmallVaraintsImgSize\textwidth,height=\SmallVaraintsImgSize\textwidth]{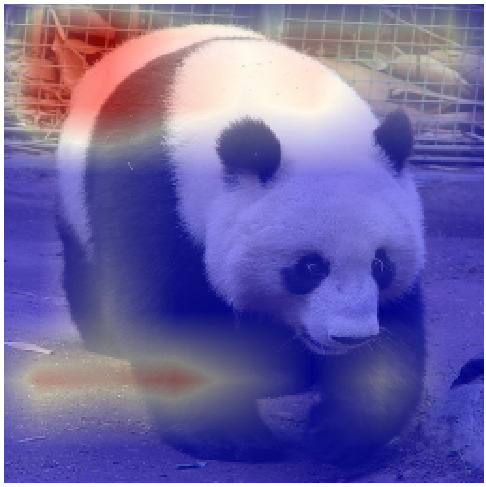} & \includegraphics[width=\SmallVaraintsImgSize\textwidth,height=\SmallVaraintsImgSize\textwidth]{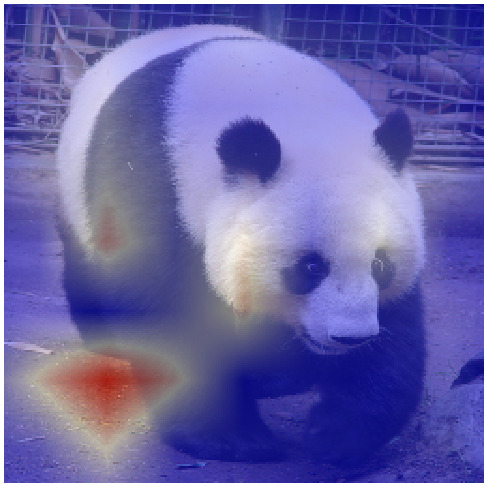} && \includegraphics[width=\SmallVaraintsImgSize\textwidth,height=\SmallVaraintsImgSize\textwidth]{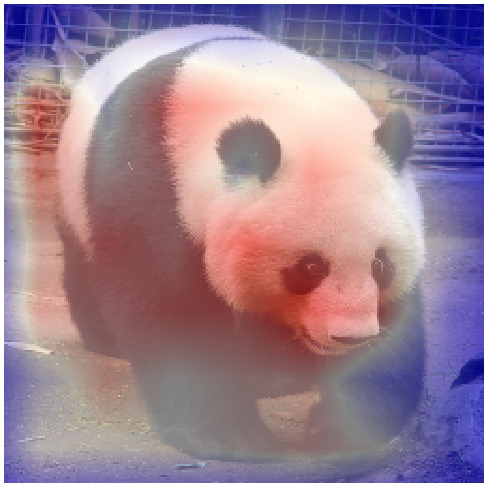} & \includegraphics[width=\SmallVaraintsImgSize\textwidth,height=\SmallVaraintsImgSize\textwidth]{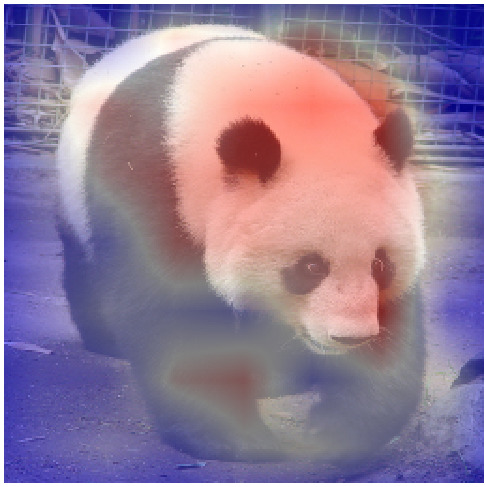} \\
	\end{tabular}
	\caption{Sanity checks~\cite{adebayo2018sanity}. First column visualizes attention using a pretrained network--nothing random. Columns two to five visualize attention when logits and weights (all-layers) are randomized. Different random initializations generate different heatmaps.}
	\label{fig:different_initialization}
\end{figure}

\noindent\underline{\textbf{Sanity Checks:}} Figure~\ref{fig:different_initialization} shows how L2-CAF is affected by randomly initialized logits-layer or weights (all layers). These sanity checks~\cite{adebayo2018sanity} emphasize a high dependency between the optimized L2-CAF (heatmap) and the network's weights.

\newcommand{\VariantsImgSize}{0.10}
\begin{figure}[t]
	\scriptsize
	\centering
	\setlength\tabcolsep{1.0pt} 
	\renewcommand{\arraystretch}{1.0}	
	\begin{tabular}{@{}cccc@{}}
		Original & L2-CAF & Softmax & Gaussian \\
		
		\includegraphics[width=\VariantsImgSize\textwidth,height=\VariantsImgSize\textwidth]{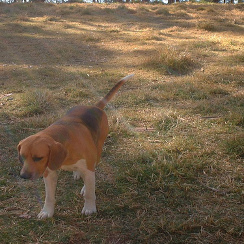} &
		\includegraphics[width=\VariantsImgSize\textwidth,height=\VariantsImgSize\textwidth]{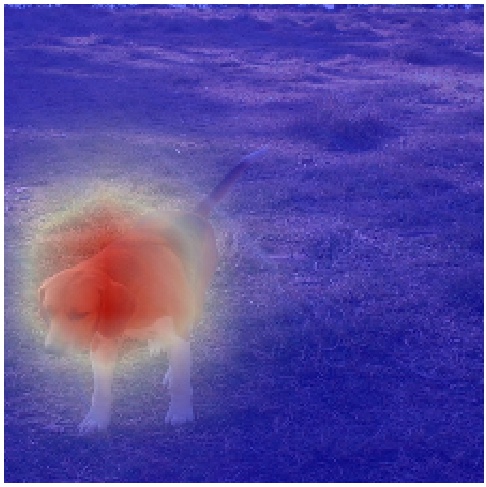} &
		\includegraphics[width=\VariantsImgSize\textwidth,height=\VariantsImgSize\textwidth]{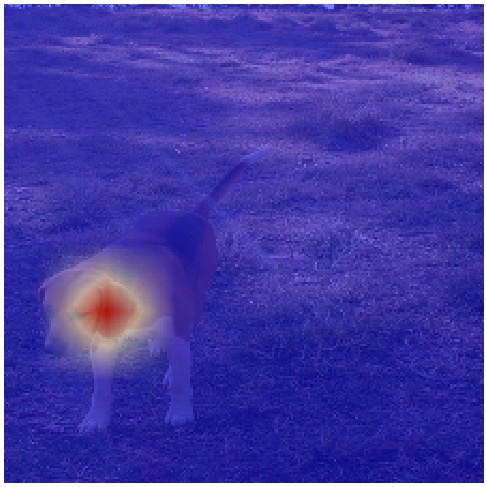} & \includegraphics[width=\VariantsImgSize\textwidth,height=\VariantsImgSize\textwidth]{dog_gauss} \\

		\includegraphics[width=\VariantsImgSize\textwidth,height=\VariantsImgSize\textwidth]{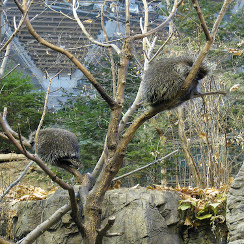} &
		\includegraphics[width=\VariantsImgSize\textwidth,height=\VariantsImgSize\textwidth]{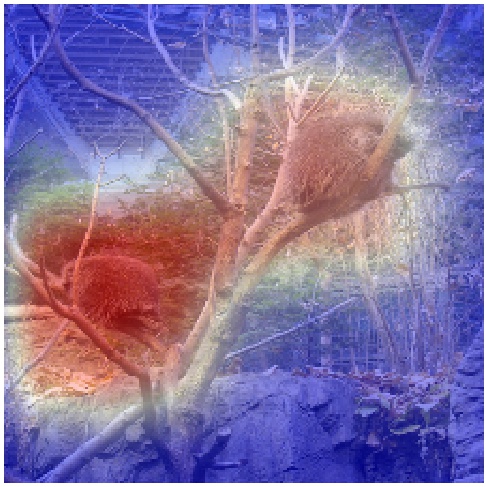} &
		\includegraphics[width=\VariantsImgSize\textwidth,height=\VariantsImgSize\textwidth]{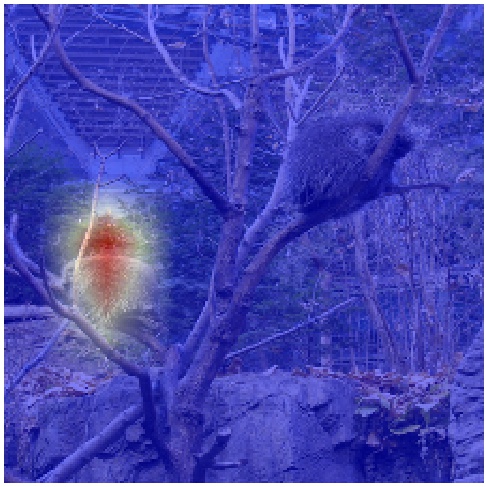} & \includegraphics[width=\VariantsImgSize\textwidth,height=\VariantsImgSize\textwidth]{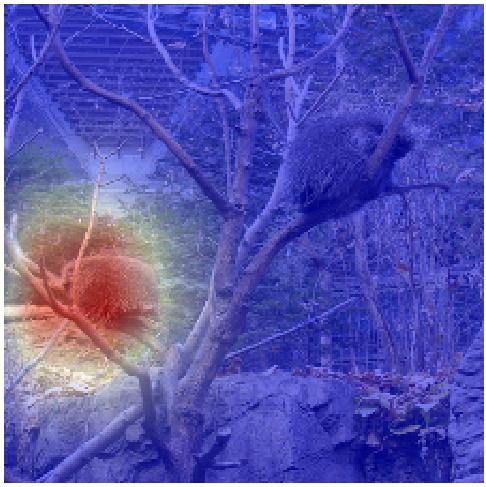} \\

		\includegraphics[width=\VariantsImgSize\textwidth,height=\VariantsImgSize\textwidth]{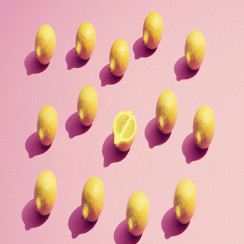} &
		\includegraphics[width=\VariantsImgSize\textwidth,height=\VariantsImgSize\textwidth]{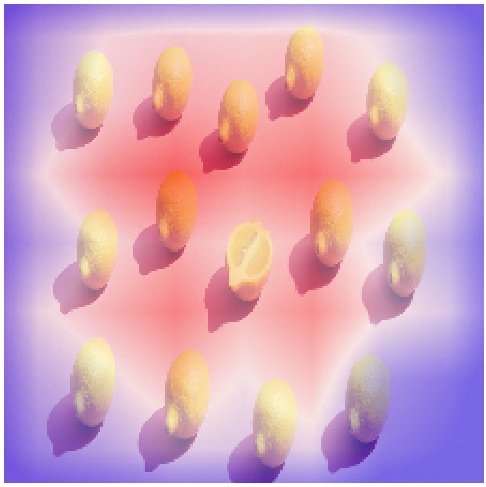} &
		\includegraphics[width=\VariantsImgSize\textwidth,height=\VariantsImgSize\textwidth]{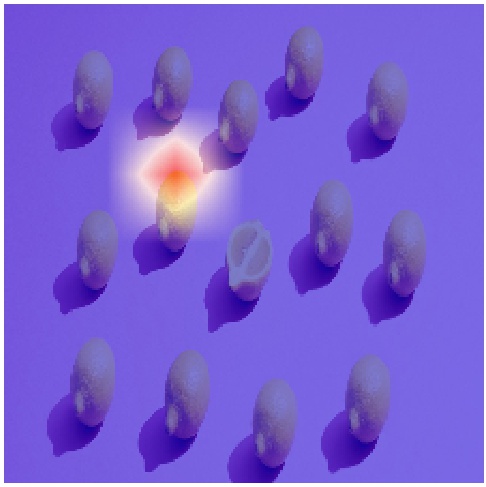} & \includegraphics[width=\VariantsImgSize\textwidth,height=\VariantsImgSize\textwidth]{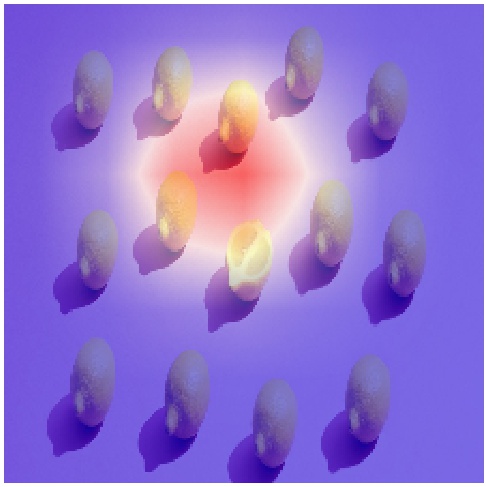} \\
	\end{tabular}
	\caption{Qualitative evaluation for alternative constraints. Softmax offers a sparse result while the Gaussian filter assumes a single mode. L2-CAF supports multi-mode. The supplementary material shows more vivid visualizations for the L2-CAF, randomly initialized, converging in slow motion.}
	\label{fig:l2_norm_variants}
\end{figure}

\noindent\underline{\textbf{Alternative Attention Constraints:}} We qualitatively compare the L2-CAF with both softmax and Gaussian constraints. These are selected for their differentiability, simplicity, and usability in recent literature. Other filtering alternatives (\eg, L1-Norm) are also feasible. Softmax is a typical attention mechanism for image captioning~\cite{xu2015show,kazemi2017show} and machine translation~\cite{vaswani2017attention}. In these problems, the softmax attention module is employed \emph{recurrently} on a single image frame or an input sentence for every output word. This fits the softmax's sparse nature. Gaussian filters have been utilized for temporal action localization~\cite{long2019gaussian,piergiovanni2018learning}. They are denser (more relaxed) compared to softmax but also assume a single mode (elliptical shape). To localize objects in images using a Gaussian constraint, we optimize the filter's  mean $\mu \in R^2$ while fixing the covariance matrix $\sigma \in R^{2\times 2}$ to the identity matrix. $\sigma$ must be constrained to avoid a degenerate solution where the Gaussian becomes a uniform distribution, \ie, $\sigma \rightarrow \infty$. All filters are optimized using the class-oblivious formulation (Sec~\ref{subsec:generic_variant}).

Figure~\ref{fig:l2_norm_variants} provides a qualitative evaluation using GoogLeNet architecture and three attention constraints. The L2-CAF identifies key region(s) for a network's output with a single glimpse. The filter prioritizes these regions quantitatively. L2-CAF supports a large spectrum of neural networks as a post-training inspection tool. It supports complex architectures including, but not limited to, encoder-decoder~\cite{kingma2013auto,rezende2014stochastic}, generative~\cite{goodfellow2014generative}, and U-shaped architectures~\cite{ronneberger2015u,li2018segmentation}. It neither undermines the performance nor raises the inference cost. L2-CAF is not the fastest attention visualization approach but is computationally cheap.

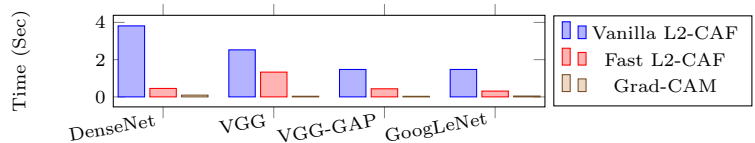
\begin{figure}[t]
	\scriptsize
	\centering
	\begin{tikzpicture} \begin{axis}[ybar, width=0.6\textwidth,height=2.8cm,enlargelimits=0.15, xtick=data, nodes near coords align={vertical},
	ylabel={Time (Sec)},
	symbolic x coords={DenseNet,VGG,VGG-GAP,GoogLeNet},
	x tick label style={rotate=10,anchor=east},
	legend style={at={(1.25,1.0)}, anchor=north,legend columns=1},
	] 
	\addplot coordinates {
		(DenseNet,3.813442469) (VGG,2.525328803) (VGG-GAP,1.472189975) (GoogLeNet,1.472507787) 
	};
	\addplot coordinates{
		(DenseNet,0.4533742428) 
		(VGG,1.329674244)
		(VGG-GAP,0.4344206333)
		(GoogLeNet,0.3109764576)
	};
	
	\addplot coordinates{
		(DenseNet,0.09335110188) 
		(VGG,0.03294756413)
		(VGG-GAP,0.0299608469)
		(GoogLeNet,0.04375116825)
	};

	\legend{Vanilla L2-CAF, Fast L2-CAF, Grad-CAM}
	\end{axis}
	\end{tikzpicture}
	\caption{Time analysis for the L2-CAF. The fast L2-CAF brings a significant speed-up while solving the same optimization problem.}
	\label{fig:time_performance}
\end{figure}

\noindent\underline{\textbf{Timing Analysis:}} Speed is the main limitation of our iterative formulation. Figure~\ref{fig:time_performance} presents a timing analysis for L2-CAF. The y-axis denotes the processing time per frame in seconds. The vanilla L2-CAF uses the default endpoints ($x$, $NT(x)$), while the fast L2-CAF uses the output of the last convolution layer and the logits $NT(x)$ as endpoints. The vanilla and fast L2-CAF are equivalent optimization problems but the fast L2-CAF provides a significant speed-up. The three fully connected layers in VGG, between the last convolution layer and the logits, limit the fast optimization technique. VGG-GAP replaces these fully connected layers with an average pooling layer, so its speed is similar to GoogLeNet. Fast L2-CAF takes $\approx 0.4$ and $0.3$ seconds per frame on VGG-GAP and GoogLeNet, respectively. The DenseNet-161's speed-up is maximum because the fast L2-CAF skips all dense blocks before the last convolution layer. 
	
	\section{Conclusion}
We have introduced the unit L2-Norm constrained attention filter (L2-CAF) as a visualization tool that works for a large spectrum of neural networks. L2-CAF neither requires fine-tuning nor imposes architectural constraints. Weakly supervised object localization is utilized for quantitative evaluation.
  State-of-the-art results are achieved on both standard and fine-tuned classification architectures.
For retrieval networks, L2-CAF significantly outperforms Grad-CAM baselines. Ablation studies highlight L2-CAF's superiority to alternative constraints and analyze L2-CAF's computational cost.


%






	\noindent\textbf{Acknowledgments:} This work was partially funded by independent grants from Office of Naval Research (N000141612713) and Facebook AI.
	\clearpage
	%
	%
	\bibliographystyle{splncs04}
	\bibliography{egbib}
	
	\clearpage
	\newcommand{\beginsupplement}{%
	\setcounter{table}{0}
	\renewcommand{\thetable}{S\arabic{table}}%
	\setcounter{figure}{0}
	\renewcommand{\thefigure}{S\arabic{figure}}%
	\setcounter{section}{0}
	\renewcommand{\thesection}{S\arabic{section}}%
	\setcounter{equation}{0}
	\renewcommand{\theequation}{S\arabic{equation}}%
}

\beginsupplement

\section{Extended Related Work}

Classification networks learn class-logits $\in R^{N_c}$. The number of logits is equal to the number of classes $N_c$. There is a clear \textit{one-to-one mapping} between classes and logits. This mapping is vital for class-activation mapping (CAM) and Grad-CAM approaches because their visualizations rely on the weights or gradients of a particular logit. In contrast, retrieval networks learn a feature embedding $\in R^{d}$. The output dimensionality does not equal the number of classes. Thus, there is no one-to-one mapping between classes and output dimensions. This lack of mapping is why CAM and Grad-CAM suffer on retrieval networks. To highlight this limitation, we train a retrieval network with various ranking losses. The following paragraphs review the two ranking losses employed in the main paper.

\subsubsection{Retrieval networks} learn a feature embedding where objects within the same class are closer than objects from different classes. To learn this feature embedding, a retrieval network is trained  with ranking losses such as contrastive, triplet, and N-pair losses. 

In the main paper, we employ triplet loss~\cite{schroff2015facenet} for its simplicity and efficiency. Equation~\ref{eq:triplet} shows the triplet loss formulation 

\begin{equation}\label{eq:triplet}
TL(a,p,n)={ { \left[  { (D(\left\lfloor a\right\rfloor,\left\lfloor p\right\rfloor)-{ D(\left\lfloor a\right\rfloor,\left\lfloor n\right\rfloor) } +m) }  \right]  }_{ + },  }
\end{equation}
where ${ \left[ \sbullet[0.75] \right]  }_{ + } = max(0,\sbullet[0.75] )$ is the hinge function and $m$ is the margin between different classes in the feature embedding. $\left\lfloor\sbullet[0.75] \right\rfloor$ and $D(,)$ are the embedding and the Euclidean distance functions, respectively. This formulation attracts an anchor image $a$ of a specific class closer to a positive image $p$ from  the same class than it is to a negative image $n$.

\begin{figure}[b]
	\centering
	\includegraphics[width=0.4\linewidth]{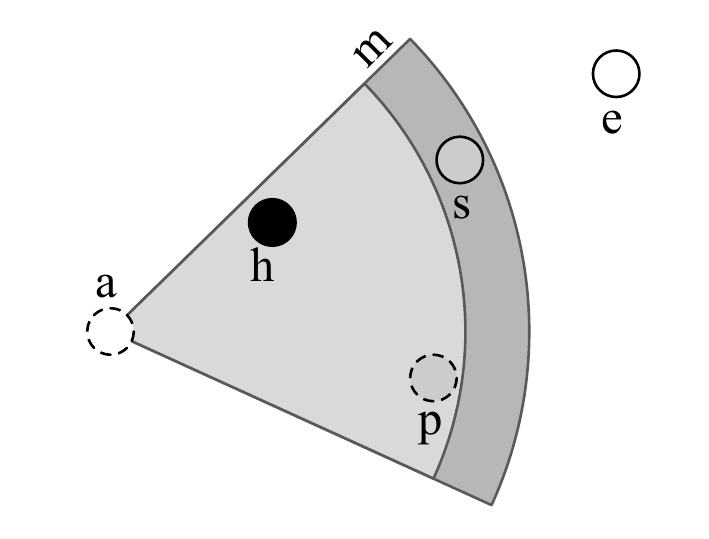}
	\caption{Triplet loss tuple (anchor, positive, negative) and margin $m$. The (h)ard, (s)emi-hard and (e)asy negatives are highlighted in black, gray, and white, respectively.}
	\label{fig:semi_neg}
\end{figure}

We leverage the semi-hard sampling~\cite{schroff2015facenet} strategy. In semi-hard negative sampling, instead of picking the hardest positive-negative samples, all anchor-positive pairs and their corresponding semi-hard negatives are considered. Semi-hard negatives are further away from the anchor than the positive exemplar, yet within the banned margin $m$ as shown in Figure~\ref{fig:semi_neg}. 

\begin{figure}[t]
	\centering
	\includegraphics[width=0.4\linewidth]{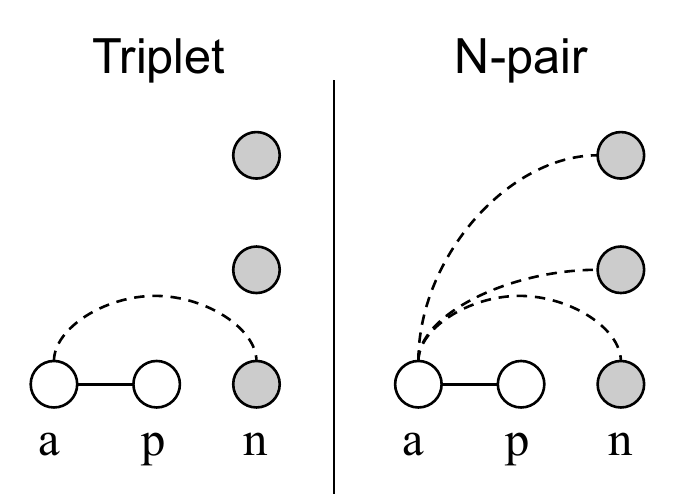}
	\caption{The difference between triplet and N-pair losses using a single positive pair $(a,p)$ and three negative ($n$) samples. The triplet loss pushes the anchor $a$ away from a selected negative sample while N-pair pushes the anchor $a$ away from all negative samples. The N-pair all-negatives approach relaxes the requirement for an efficient negative mining strategy.}
	\label{fig:triplet_vs_npair}
\end{figure}

The performance of triplet loss relies heavily on the sampling strategy because every anchor sample is paired with a single negative sample. N-pair loss mitigates this limitation by pairing every anchor with all negative samples within a mini-batch. Figure~\ref{fig:triplet_vs_npair} depicts the difference between triplet and N-pair losses. Equation~\ref{eq:n_pair} shows the N-pair loss formulation 
\begin{equation}\label{eq:n_pair}
\text{NPL} = -\log{\frac{exp(\lfloor a \rfloor \lfloor p \rfloor)}{exp(\lfloor a \rfloor \lfloor p \rfloor)+\sum_{n\in B}^{}exp(\lfloor a \rfloor \lfloor n \rfloor)}},
\end{equation}
For N-pair loss, a training batch contains a single positive pair from each class. Thus, a mini-batch will have $b/2$ positive pairs and every anchor is paired with $b-2$ negatives, where $b$ is the mini-batch size.

\subsubsection{Weakly supervised object localization (WSOL)} approaches localize objects inside images using the class label only. Attention visualization approaches (\eg, CAM) generate class-specific attention heatmaps. A simple segmentation of the heatmap provides a localization bounding box. Attention-based approaches do not require bounding box annotations during training. Thus, these approaches reduce the cost of data annotation; yet, they tend to localize the most discriminative part of an object, not the entire object. For instance, an attention-based approach would focus on the cat's head and ignore other parts such as legs. Thus, the result bounding box partially covers the object (\eg, cat's head) while it should cover all its parts.

Attention-based approaches focus on the most discriminative part because classification CNNs focus on the most discriminative features to boost their classification performance. To mitigate this limitation, Choe and Shim~\cite{choe2019attention} proposed an \textbf{a}ttention-based \textbf{d}ropout \textbf{l}ayer (ADL) while Zhang~\etal~\cite{zhang2018adversarial} proposed \textbf{a}dversarial \textbf{co}mplementary \textbf{l}earning (ACoL). Both approaches have the same core objective,~\ie, hide the most discriminative feature (\eg, cat's head feature) so the classifier identifies less discriminative parts. The following paragraphs review ACoL and ADL.

\begin{figure}[t]
	\centering
	\scriptsize
	\includegraphics[width=0.8\textwidth]{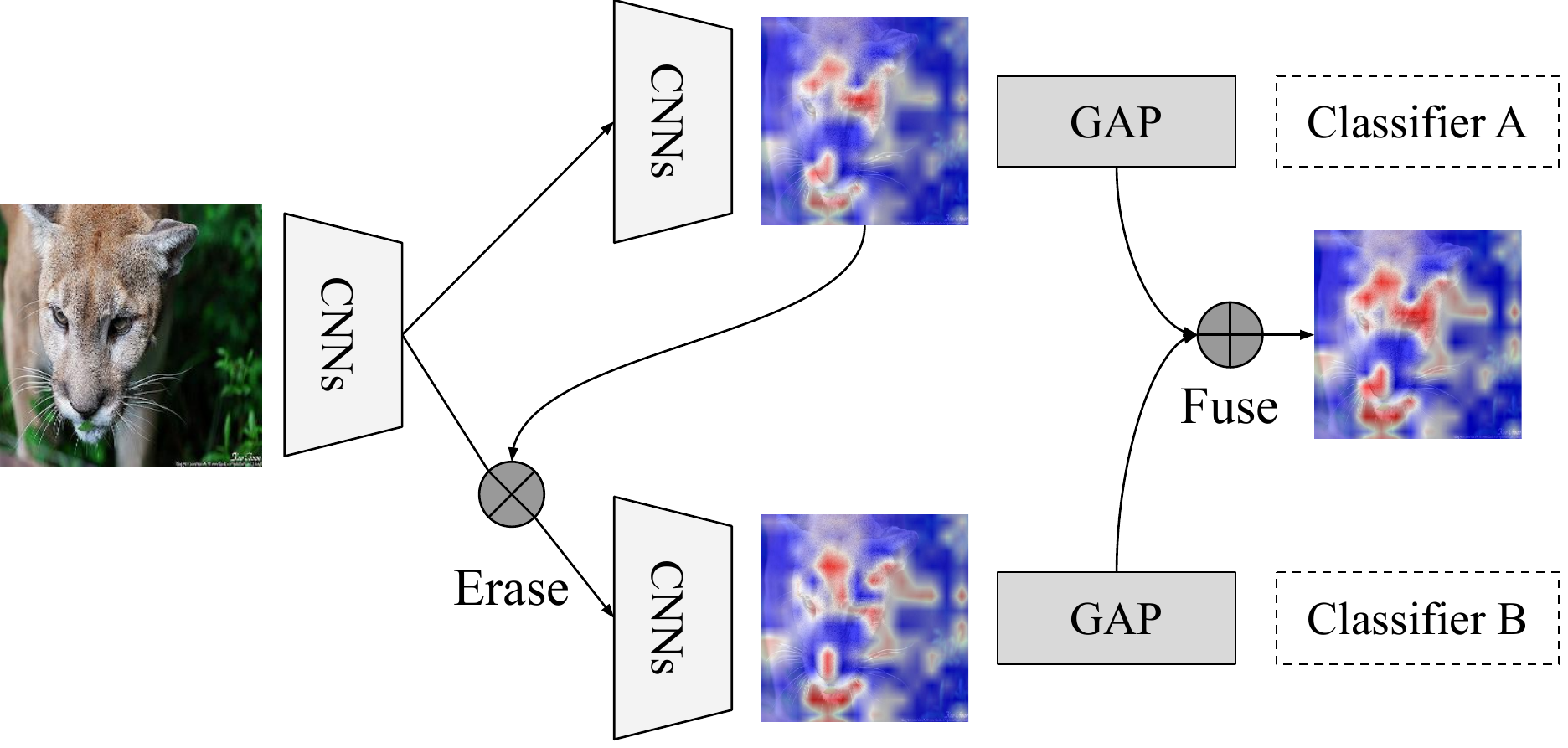}
	\caption{An illustration of the ACoL method; A classification network is trained with two complementary classifier heads ($A$ and $B$). Classifier $A$ is presented with a localization map that highlights the most discriminative parts. The discriminative-parts' features are erased from the input features of classifier $B$. Accordingly, classifier $B$ learns complementary parts of an object. GAP refers to global average pooling.}
	\label{fig:acol}
\end{figure}

Zhang~\etal~\cite{zhang2018adversarial} train a classification network with two classification heads ($A$ and $B$). During training, the localization heatmap for classifier $A$ is obtained. This localization heatmap identifies the most discriminative region. Zhang~\etal~\cite{zhang2018adversarial} use this heatmap to guide an erasing operation on the intermediate feature maps of classifier $B$. This drives classifier $B$ to discover complementary object-related regions. Thus, the two classifiers are trained to exploit complementary object regions and obtain integral object localization. Figure~\ref{fig:acol} depicts an illustration for this training strategy.

To eliminate the auxiliary classification head in ACoL, Choe and Shim~\cite{choe2019attention} proposed an attention-based dropout layer (ADL). Similar to ACoL~\cite{zhang2018adversarial}, ADL obtains a localization heatmap during training. From the heatmap, ADL produces both a drop-mask and an importance-map through simple-thresholding and sigmoid-activation, respectively. Applying the drop-mask drives the model to learn the less discriminative parts, which improves the localization performance. In contrast, applying the importance-map highlights the most discriminative region which improves the classification performance. During training, either the drop-mask or the importance-map is stochastically selected at each iteration, and then the selected one is applied to the input feature map through a spatialwise multiplication as shown in the next Figure~\ref{fig:adl}


\begin{figure}[t]
	\centering
	\scriptsize
	\includegraphics[width=1.0\textwidth]{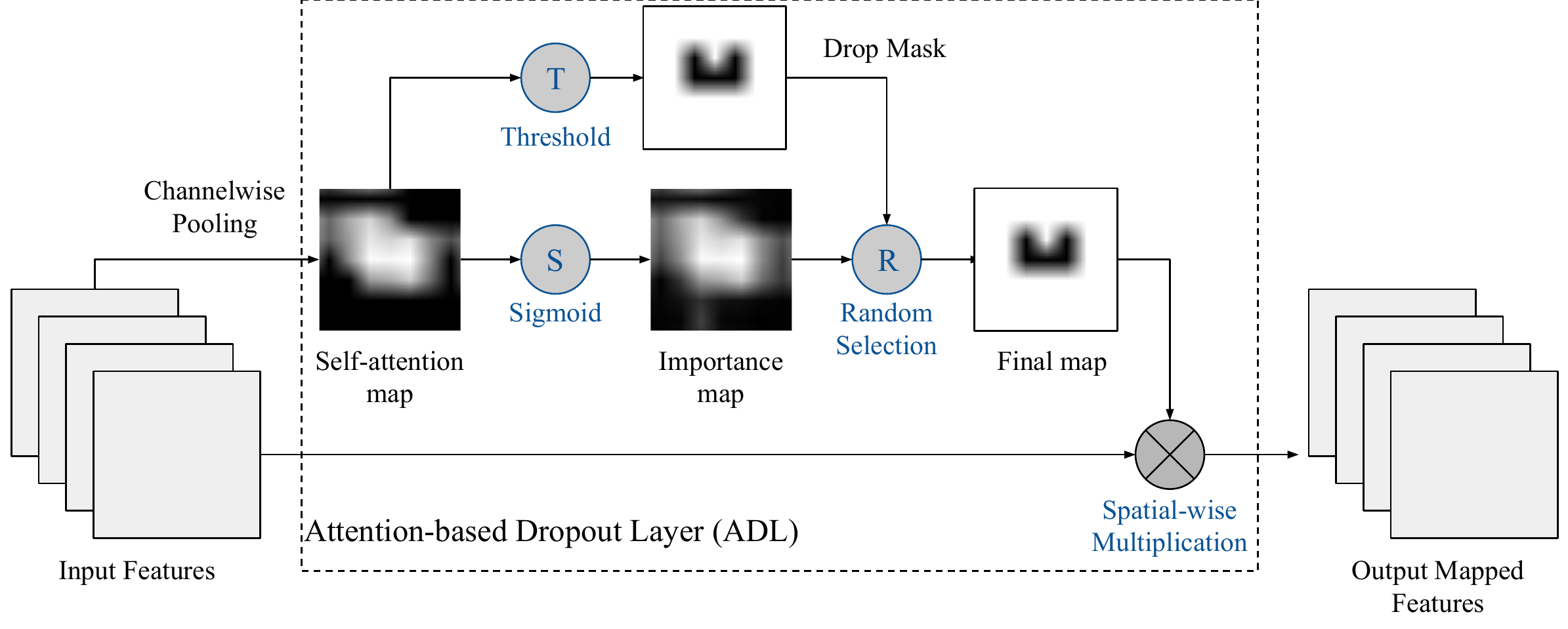}
	\caption{ADL illustration diagram. The self-attention map is generated by channelwise average pooling of the input feature map. Based on the self-attention map, a drop-mask is produced by thresholding and an importance-map is produced by a sigmoid activation. At every training iteration, either the drop-mask or the importance-map is selected and applied to the input feature map.}
	\label{fig:adl}
\end{figure}
\section{Extended Experiments}

\newcommand{\sanityCheckImgSize}{0.20}

\begin{figure}[t]
	\centering
	\tiny
	\setlength\tabcolsep{1.0pt} 
	\renewcommand{\arraystretch}{1.0}	
	\begin{tabular}{@{}ccc@{}}
		Pretrained & Random Logits & Random Weights
		\\ 
		
		\includegraphics[width=\sanityCheckImgSize\textwidth,height=\sanityCheckImgSize\textwidth]{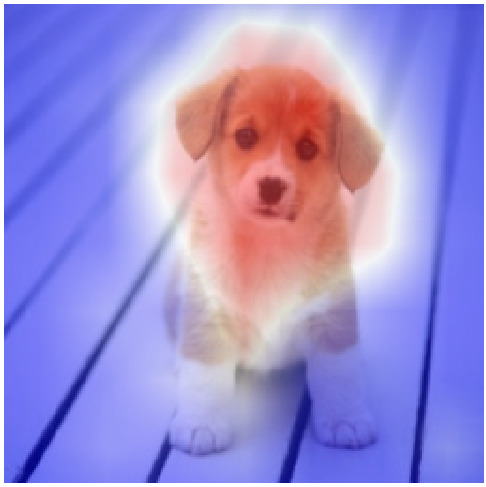} &
		\includegraphics[width=\sanityCheckImgSize\textwidth,height=\sanityCheckImgSize\textwidth]{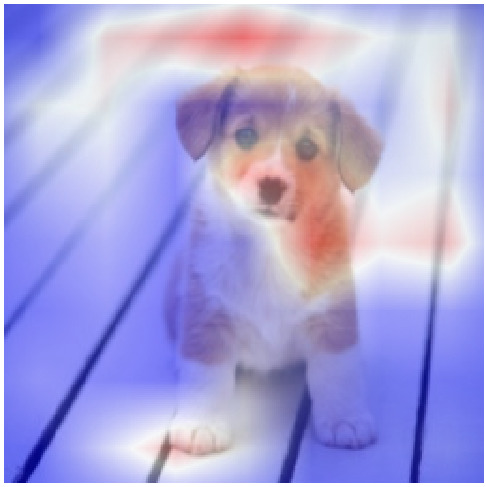} & \includegraphics[width=\sanityCheckImgSize\textwidth,height=\sanityCheckImgSize\textwidth]{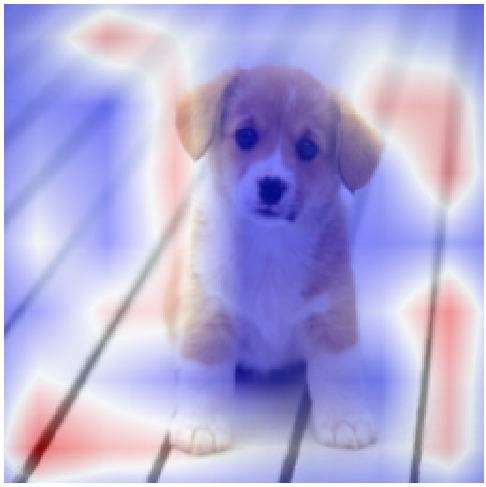} \\
		
		\includegraphics[width=\sanityCheckImgSize\textwidth,height=\sanityCheckImgSize\textwidth]{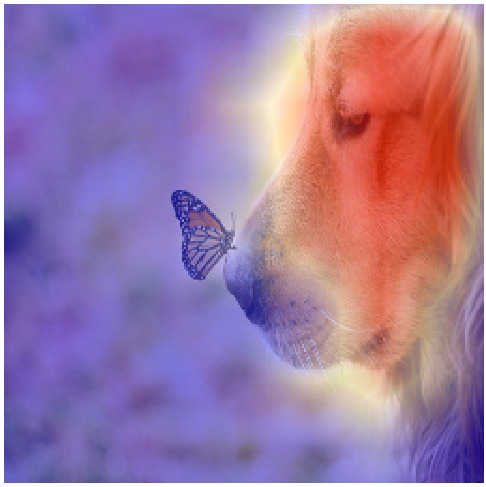} &
		\includegraphics[width=\sanityCheckImgSize\textwidth,height=\sanityCheckImgSize\textwidth]{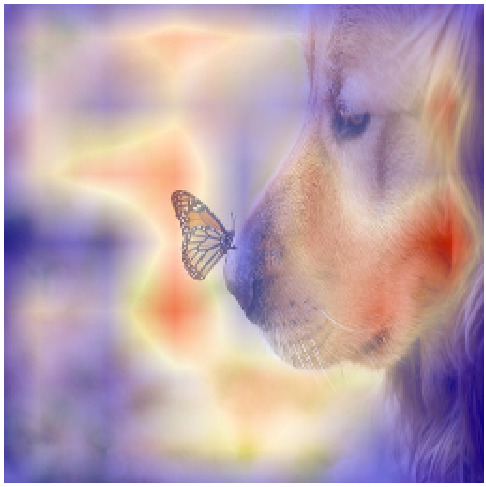} & \includegraphics[width=\sanityCheckImgSize\textwidth,height=\sanityCheckImgSize\textwidth]{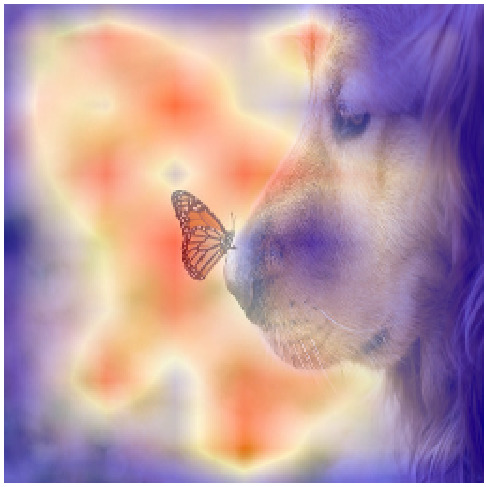} \\
		
		\includegraphics[width=\sanityCheckImgSize\textwidth,height=\sanityCheckImgSize\textwidth]{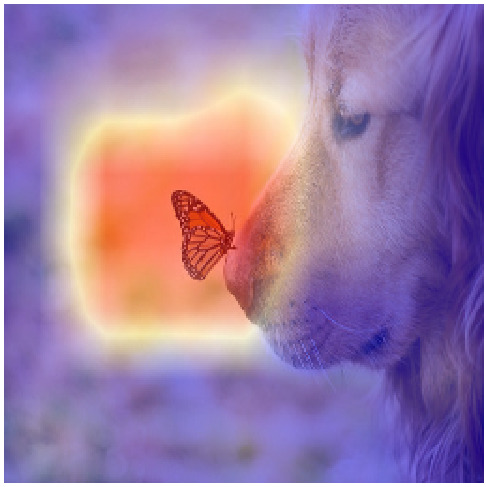} &
		\includegraphics[width=\sanityCheckImgSize\textwidth,height=\sanityCheckImgSize\textwidth]{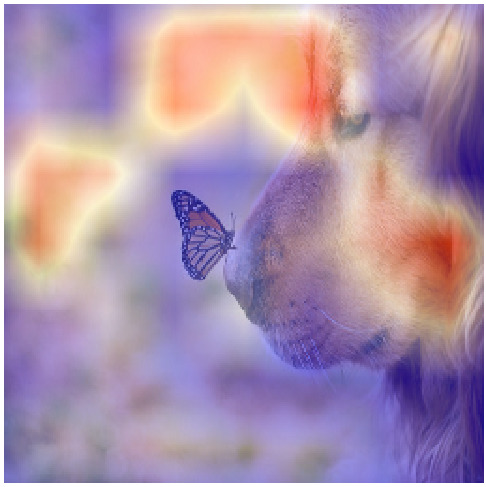} & \includegraphics[width=\sanityCheckImgSize\textwidth,height=\sanityCheckImgSize\textwidth]{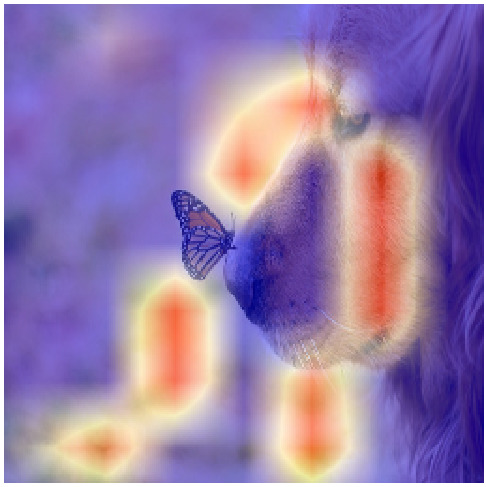} \\
	\end{tabular}
	\caption{First column depicts attention using a pretrained network--nothing random. Second and third columns depict attention when logits and weights (all-layers) are randomized.}
	\label{fig:sanity_checks}
\end{figure}

\noindent\textbf{Implementation Details For Retrieval Networks:} To evaluate the localization performance quantitatively, we leverage both triplet~\cite{schroff2015facenet} and N-pair~\cite{sohn2016improved} ranking losses. We use the default settings for each loss; the N-pair's embedding is unnormalized while the triplet loss's embedding is normalized to the unit-circle and a margin $m=0.2$ is utilized. We employ ResNet-50~\cite{he2016deep} and GoogLeNet~\cite{szegedy2015going} as backbones.  These are standard architectures for evaluating ranking losses~\cite{sohn2016improved,wang2017deep,oh2016deep}. Both architectures are trained for 5K iterations. VGG architecture is omitted because it overfits on these datasets.  Similar to Hermans~\etal~\cite{hermans2017defense}, the last convolution layer is followed by a global average pooling layer then a single fully connected layer,~\ie, a feature embedding $\in R^{128}$.

\noindent\textbf{Evaluation metrics:} \underline{For retrieval}, we utilize both Recall@1 (R@1) and the Normalized Mutual Information (NMI) metrics. NMI score $\in [0,1]$ measures the agreement between the true and predicted cluster assignments. $\text{NMI}=\frac { I(\Omega ,C) }{ \sqrt { H(\Omega )H(C) }  } ,$ where $\Omega =\{\omega_1,..,\omega_n\}$ is the ground-truth clustering while $C=\{c_1,...c_n\}$ is a clustering assignment for the learned embedding. $I(\sbullet[0.5],\sbullet[0.5])$ and $H(\sbullet[0.5])$ denote mutual information and entropy, respectively. We use K-means to compute $C$. \underline{For localization}, we follow the same evaluation procedure in \cite{zhou2016learning,selvaraju2017grad} for classification networks. We replace the top-1 by R@1 metric to decide if the network's output is correct or not. The same IoU $> 50\%$ criterion is used to evaluate localization.

Figures~\ref{fig:sanity_checks} and~\ref{fig:different_initialization} show how random initialization for the logit-layer, or the whole network, affects attention visualization. These sanity checks~\cite{adebayo2018sanity} emphasize a high dependency between the proposed L2-CAF and the weights of the network.

\begin{figure}[h]
	\centering
	\tiny
	\setlength\tabcolsep{3.0pt} 
	\renewcommand{\arraystretch}{1.0}	
	\begin{tabular}{@{}c cccc@{}}
		Pretrained &\phantom{ab}& \multicolumn{3}{c}{Different Random Logits Initializations} 
		\\ 
		\includegraphics[width=\sanityCheckImgSize\textwidth,height=\sanityCheckImgSize\textwidth]{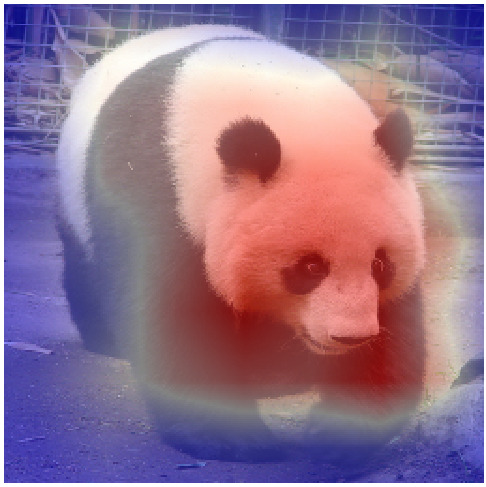} &&
		\includegraphics[width=\sanityCheckImgSize\textwidth,height=\sanityCheckImgSize\textwidth]{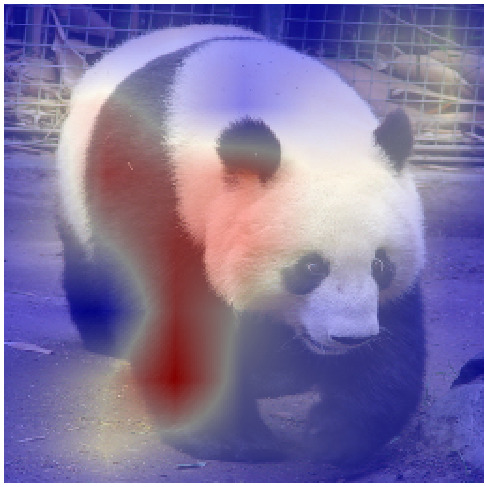} & \includegraphics[width=\sanityCheckImgSize\textwidth,height=\sanityCheckImgSize\textwidth]{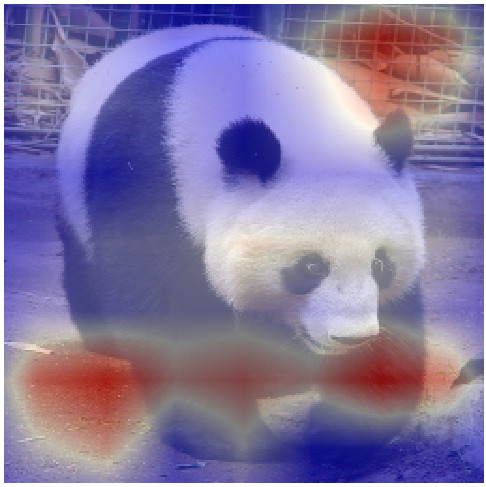} & \includegraphics[width=\sanityCheckImgSize\textwidth,height=\sanityCheckImgSize\textwidth]{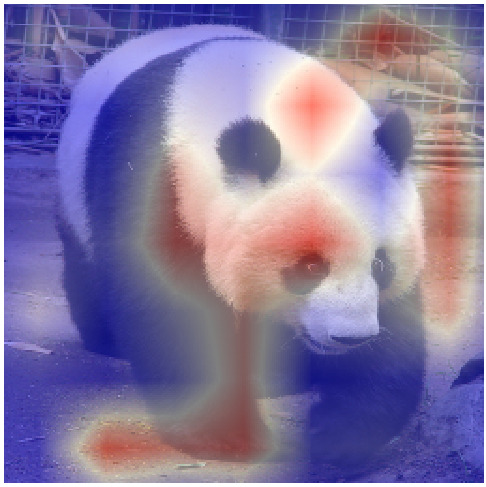} \\
		
	\end{tabular}
	\caption{Different random logit initializations, columns 2-4, generate different heatmaps.}
	\label{fig:different_initialization}
\end{figure}
\end{document}